\documentclass[sigconf, nonacm]{acmart}

\usepackage{xspace}
\usepackage{xcolor}
\usepackage{enumitem}
\usepackage{caption}
\usepackage{subcaption}

\newcommand{\system}{\texttt{F\lowercase{ood}SG}\xspace}
\newcommand{\dmapp}{\texttt{JHFoodlg}\xspace}
\newcommand{\dentalapp}{\texttt{eDental}\xspace}
\newcommand{\app}{\texttt{F\lowercase{ood(lg)}}\xspace}

\newcommand{\scl}{\texttt{SCL}\xspace}
\newcommand{\model}{\texttt{FoodSG-SCL}\xspace}
\newcommand{\dataset}{\texttt{FoodSG-233}\xspace}

\newcommand{\ind}[1]{{#1}}

\begin{document}
\title{
From Plate to Prevention: A Dietary Nutrient-aided Platform for Health Promotion in Singapore}

\author{
Kaiping Zheng$^1$, Thao Nguyen$^1$, Jesslyn Hwei Sing Chong$^2$, Charlene Enhui Goh$^3$\\ Melanie Herschel$^4$, Hee Hoon Lee$^5$, Changshuo Liu$^1$, Beng Chin Ooi$^1$, Wei Wang$^6$, James Yip$^7$
}
\affiliation{
\country{
\vspace{2mm}
$^1$School of Computing, National University of Singapore \\
$^2$Dietetics and Nutrition Department, Ng Teng Fong General Hospital\\
$^3$Faculty of Dentistry, National University of Singapore \\
$^4$Universität Stuttgart, Germany\\
$^5$ Allied Health Division, Ng Teng Fong General Hospital \\
$^6$TikTok Singapore \\
$^7$National University Heart Centre, Singapore}
}
\renewcommand{\shortauthors}{Kaiping Zheng, Thao Nguyen, Jesslyn Hwei Sing Chong, Charlene Enhui Goh, Melanie Herschel, Hee Hoon Lee, Changshuo Liu, Beng Chin Ooi, Wei Wang, and James Yip}

\begin{abstract}
\ind{Singapore has been striving to improve the provision of healthcare services to her people. In this course, the government has taken note of the deficiency in regulating and supervising people's nutrient intake, which is identified as a contributing factor to the development of chronic diseases. Consequently, this issue has garnered significant attention.
In this paper, we share our experience in addressing this issue and attaining medical-grade nutrient intake information to benefit Singaporeans in different aspects.
To this end, we develop the \system platform to incubate diverse healthcare-oriented applications as a service in Singapore, taking into account their shared requirements.
We further identify the profound meaning of localized food datasets
and systematically clean and curate a localized Singaporean food dataset \dataset.
To overcome the hurdle in recognition performance brought by Singaporean multifarious food dishes, we propose to integrate supervised contrastive learning into our food recognition model \model for the intrinsic capability to mine hard positive/negative samples and therefore boost the accuracy. 
Through a comprehensive evaluation, we present
performance results of the proposed model and insights on
food-related healthcare applications.
The \dataset dataset has been released in \url{https://foodlg.comp.nus.edu.sg/}.}
\end{abstract}

\maketitle
\pagestyle{plain}

\section{Introduction}

\ind{
Singapore has been making persistent efforts to deliver superior healthcare, with a top priority on enhancing the quality of life for her people. 
Drawing upon considerable support for medical research and innovative technology development, the research community of healthcare analytics in Singapore has attained significant achievements in effectively integrating multifarious types of healthcare data pertaining to each individual, and therefore empowering high-quality data-driven healthcare delivery. 
As data management technologies continue to advance, electronic medical records (diagnoses, medications, lab tests, among others) have been integrated from all levels of medical institutions (hospitals, clinics, pharmacies). In addition, sensor data, which records time-series vital signs through wearable devices, has been collected as well, enabling the acquisition of heterogeneous healthcare data. The collected healthcare data has been utilized for different healthcare applications, such as chronic disease progression modeling~\cite{zheng2017resolving, zheng2017capturing, zheng2021pace}, and disease diagnosis~\cite{zheng2020tracer, zheng2022dyhealth}.
Consequently, Singapore has expanded the capacity of the healthcare system for reacting to the people who are sick, spanning primary, acute, and long-term care, with the target of building a comprehensive healthcare data profile for each individual. Such measures reflect the improved healthcare system functioning, evidenced by reduced hospital readmission rates and Singapore's relatively high life expectancy compared to other nations worldwide.

The swift development of Singapore's national power, facilitated by the robust measures of government administration, has greatly improved Singaporeans' living standards and diet quality. The substantial material prosperity has largely changed people's dietary behaviors, thus bringing about new medical challenges~\cite{bee2022singapore}. 
Without adequate regulation and supervision of the nutrient intake, e.g., sugar and saturated fat, in people's diets, the risk of developing chronic diseases such as diabetes, hypertension, and hyperlipidemia could be increased, leading to a significant economic burden on Singaporeans~\cite{min2019survey, achananuparp2018does}.

Singapore recognized this challenge as early as 2016 and responded by implementing several key initiatives. In April of that year, the Minister of Health (MOH) called for a whole-of-nation effort to combat diabetes, with the aim of raising public awareness of chronic disease management among Singaporeans~\cite{bee2022singapore, ow2021war}. Afterward, diabetes (hyperglycemia) is jointly taken into consideration with hypertension (high blood pressure) and hyperlipidemia (high cholesterol), i.e., 3H problems, which are identified as a grand challenge in Singapore, necessitating timely control via innovative artificial intelligence solutions~\cite{AI_in_Health_for_3H}.
We formed a diversified team of clinicians, dietitians, dentists, and computer scientists in 2017 to tackle this challenge. We note that a successful solution for combating such chronic diseases highly depends on the medical-grade nutrient intake data, which is not readily available in existing healthcare data. 
This is due to the impracticality of requesting each individual to log their food intake for every meal and the complexity of calculating nutrient intake, which is not easily learnable~\cite{thompson2010need}.

Fortunately, benefiting from the advancement of computer vision and the popularization of mobile electronic devices, it becomes a viable solution to record nutrient intake information by prompting users to snap photos of their consumed food, utilizing food recognition techniques to identify food types from images, and assessing diets to provide much-needed nutrient intake information~\cite{min2019survey}. Nonetheless, it is still challenging to attain accurate, medical-grade nutrient intake information that could facilitate the monitoring of individuals' dietary consumption and trigger appropriate interventions if necessary.

The most paramount challenge we encounter is ensuring the usability of our solution. We investigate the local situations in Singapore and sort out the potential application scenarios wherein accurate nutrient intake information could be effectively leveraged. Serving as a cornerstone to support a range of diverse scenarios, our proposal should satisfy the shared requirements and is thus versatile and flexible for deployment.
Taking a step further, influenced by its multi-ethnic culture and citizen population, Singaporean cuisine is highly diverse~\cite{sgfood}. Such uniqueness renders it imperative to collect a representative dataset encompassing different localized Singaporean dishes.
In addition, although food recognition is a promising technique for retrieving nutrient intake data from users' captured food photos, how to achieve accurate recognition results is challenging. Two critical issues, i.e., intra-class variation and inter-class resemblance, which are prevalent in real-world food images, result in the existence of hard positive samples and hard negative samples, respectively. Both issues hinder us from achieving effective food recognition and hence accurate nutrient intake retrieval.

To overcome these barriers, we built a systematic solution \system in 2017, which was initially developed with the primary objective of aiding individuals in managing 3H problems.
With the further development of the healthcare industry, Singapore has escalated its focus on monitoring and regulating people's nutrient intake to foster a culture of healthy living.
In September 2022, MOH initiated ``Healthier SG''~\cite{healthiersg}  to establish an ecosystem of support for better health. This initiative seeks to transform the healthcare system from caring for patients reactively to preventing people from falling sick proactively, which entails advocating advanced health-seeking behaviors and healthier lifestyles.
Consequently, \system has served for diverse significant scenarios in Singapore over the years as a fundamental medical facility to collect medical-grade nutrient intake information for each individual, which benefits Singaporeans in every aspect of life.

Developed thus far, powered by \system, our accomplishments include: (i) assisting in preventing the enrolled prediabetic patients from progressing into diabetes in a 3H prevention program, (ii) facilitating the co-management of patients and their dentists on oral health management, (iii) planning athletes' diets with suggested nutrients, and (iv) releasing a public app for general users to monitor their nutrient intake themselves. 
With such broad applicability, we bring benefits to Singaporeans in different aspects, promote lifestyle adjustments and finally, help them maintain healthy physical and mental well-being.

In this paper, we share our experience in this process with researchers and practitioners in the data management community.
Specifically, we have established \system as a full-fledged platform since 2017 with synergistic collaboration and expertise from different disciplines ranging from dietitians and clinicians to data scientists and system architects in our team. \system is general and flexible to satisfy the requirements of diverse application scenarios through effective dietary nutrient intake retrieval and analysis. 
At the core of \system, both a high-quality localized Singaporean food dataset and an accurate food recognition model are crucial. 
For the former, we have systematically collected and curated the \dataset dataset, while for the latter, we propose to integrate supervised contrastive learning (\scl) which is beneficial for learning from hard positive/negative samples.
We summarize our main contributions below.

\begin{itemize}[leftmargin=*]
	\setlength\itemsep{1mm}
	\item Stemming from the close collaboration with different healthcare sectors in Singapore, we take into account the nutrient intake information from consumed food data, bridging this gap towards building a comprehensive healthcare data profile for each individual in Singapore. 
    Consequently, we bring tangible and sustainable benefits to Singaporeans in multiple aspects, spanning from chronic disease prevention, and oral health management, to healthy lifestyle transformation.
	
	\item We design and develop the \system platform to retrieve medical-grade nutrient intake information, serving a variety of healthcare-oriented applications and healthy living programs in Singapore. We further emphasize the importance of localized datasets to the healthcare management of each specific country, and collect and clean a localized Singaporean food dataset \dataset with a systematic curation pipeline. To achieve effective food recognition for accurate nutrient intake retrieval, we propose to integrate \scl into the model design and devise the \model model to facilitate the mining of hard positive/negative samples caused by intra-class variation and inter-class resemblance.
	
	\item We share our experience in the process of solving this enormous nationwide challenge to retrieve medical-grade nutrient intake data and make the \dataset publicly available to foster data management research in food computing. In addition, with an extensive experimental evaluation, we validate the effectiveness of \model, and also provide the database community with fresh insights into real-world healthcare applications.
	
\end{itemize}
}

\ind{The remainder of this paper is structured as follows. We introduce necessary preliminaries, including background and challenges in Section~\ref{sec:preliminaries}. We elaborate on \system as a dietary nutrient-aided healthcare platform, covering the rationale of our solution, the overview of \system, and its supported representative application scenarios in Singapore in Section~\ref{sec:foodsg}. We further describe our released localized Singaporean food dataset \dataset, and our proposed \scl-based effective food recognition model \model in Section~\ref{sec:data collection and curation} and Section~\ref{sec:model}, respectively. The experimental evaluation is discussed in Section~\ref{sec:experiments} and the related work is reviewed in Section~\ref{sec:related work}. Finally, we conclude in Section~\ref{sec:conclusions}.}

\vspace{0mm}
\section{\ind{Preliminaries}}
\label{sec:preliminaries}

\subsection{\ind{Background}}
\ind{
	
Nutrient intakes, such as 
saturated fat, sodium, sugar, and carbohydrate, exert immense influence
on people's health conditions. For instance, nutritional inadequacy and imbalance due to improper dietary intake tend to increase the risk of people developing certain diseases such as diabetes, hypertension, and hyperlipidemia~\cite{tucker2016nutrient, gruchow1985alcohol, oza2009trends}. Therefore, serving as a modifiable risk factor, nutrient intake could be monitored and regulated to reduce health hazards for people. 

Despite the vital significance of the nutrient intake information, it is non-trivial to collect such data in an accurate and medically useful manner as supplementary information to people's healthcare data, facilitating diverse downstream analytic applications. 
One main reason lies in the notoriously tedious logging and quantification of
daily nutrient intake, which renders people less motivated to monitor their diets and hence, the quality of collected nutrient data degraded~\cite{thompson2010need}. Benefiting from the advancement of computer vision, food recognition emerges as a promising direction for analyzing users' snapped food photos, predicting the corresponding food type, and retrieving nutrient intake data automatically.

Nonetheless, this is still challenging in practice as people's daily food intake is generally of high variability in terms of types, portions, etc. Fortunately, the particular dietary behaviors of Singaporean people offer advantages to alleviate this issue, i.e., a significant percentage of daily diets are ``standardized''. As a matter of culture, Singaporeans purchase foods in hawker centers (or cooked food centers)~\cite{hawkercentre, tan2020we} 
of which the locations, menus, and prices are regulated by the government. For most Singaporeans' dining, despite the varied choices, the consumed food can be largely regarded as a finite set of items from the data collection perspective. This facilitates effective nutrient intake retrieval via self-snapped photos.
In addition, for other application scenarios in healthcare institutions, such as hospital wards, sports institutes, and rehabilitation centers, patients are generally provided with predefined set meals, which brings down the possible variety of daily diets and thereby eases nutrient intake retrieval.

The factors above largely reduce the difficulty of establishing an accurate mapping between food categories and the corresponding nutrients for intake monitoring and diet assessment. However, this still needs to be grounded on the availability of an official mapping between standardized menus and nutrient information based on local situations.
We address this by incorporating the standard nutritional guidelines and food composition from the Singapore Health Promotion Board (HPB) with dietitians' continuous consolidation. 
}

\begin{figure}[t]
	\centering
	\includegraphics[width=\linewidth]{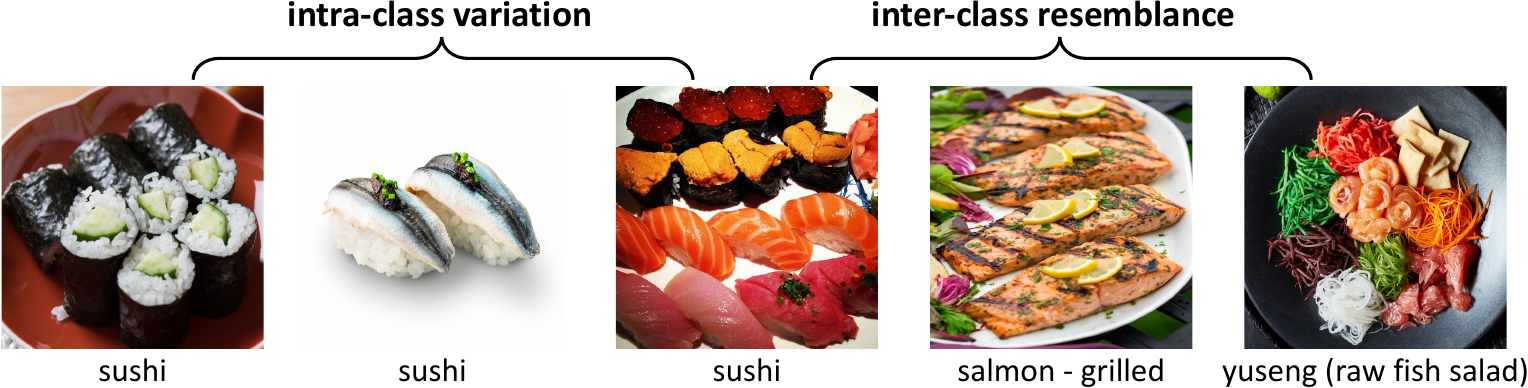}
	\caption{Two common issues in real-world food datasets.}
	\label{fig:data issues}
\end{figure}

\subsection{\ind{Challenges}}
\label{subsec:challenges}

\ind{
We have thus far discussed how we overcome the preliminary barriers towards the goal of accurate nutrient intake data retrieval from people's daily diets in Singapore. We next dive into the three fundamental challenges in attaining this goal further.

\vspace{1mm}
\noindent
\textbf{Usability.}
Since nutrient intake retrieval from daily diets serves as a cornerstone for analyzing people's health conditions, the proposed solution should be highly robust and usable to support different application scenarios. Therefore, it is highly desired to abstract the common requirements from the various scenarios and propose a versatile and flexible solution to suit diverse environments.

\vspace{1mm}
\noindent
\textbf{Uniqueness.} Singaporean cuisine is renowned for its rich blend of dishes from different countries, including Malay, Indonesian, Chinese, Indian, Peranakan, Western, Japanese, Korean, and Thai, among others~\cite{sgfood}. Such food diversity poses an enormous challenge to effective food recognition for accurate nutrient intake retrieval in Singapore, and leads to the uniqueness of food recognition in Singaporean food. Prior studies either make use of hand-crafted features for food recognition~\cite{chen2009pittsburgh, yang2010food, anhimopoulos2014food} or more recently turn to deep neural network models for boosted performance~\cite{kagaya2014food, wu2016learning}, 
fueling the releasing of public food datasets~\cite{bossard2014food, chen2016deep, matsuda2012multiple, kawano2014automatic} in the meanwhile. Although these datasets are useful in numerous studies, they lay emphasis on general food types or different geographical regions and hence, may not be applicable to more specific areas, in our case, Singapore. To power effective food recognition for Singaporean cuisine, it is imperative to have access to a comprehensive localized Singaporean dataset for analysis and development.

\vspace{1mm}
\noindent
\textbf{Efficacy.}
There are two critical issues prevalent in real-world food datasets exemplified in Figure~\ref{fig:data issues}: (i) intra-class variation meaning that the food images in the same food category may exhibit high dissimilarity in appearance, e.g., varied types of ``sushi'', which gives rise to hard positive samples, and (ii) inter-class resemblance indicating that some dishes may look similar to each other despite being in different categories, such as ``sushi'' and ``salmon - grilled'', which results in hard negative samples. The existence of such hard positive/negative samples tends to impede the food recognition approaches to achieve satisfactory performance. Therefore, advanced techniques should be introduced to alleviate both issues to improve the efficacy of food recognition.
} 

\vspace{-2mm}
\section{\ind{\system}}
\label{sec:foodsg}

\ind{In this section, we present our proposed dietary nutrient-aided healthcare platform \system. We start with how \system addresses each respective challenge in Section~\ref{subsec:challenges} and then demonstrate its architecture overview. Finally, we elaborate on how \system serves the four representative application scenarios in Singapore.}

\subsection{\ind{Our Solution}}

\ind{
Our solution to address the fundamental challenges for nutrient intake retrieval is three-fold.

Firstly, we identify the shared requirements in our close collaborations with different healthcare sectors, clinicians, and dietitians on food recognition-based applications. We then design and establish \system, a full-fledged platform that supports diverse healthcare applications in Singapore driven by effective food recognition and dietary nutrient analysis. With the \system platform as a service, we manage to bring huge benefits to health promotion in Singapore from various aspects, including 3H 
prevention, dental care management, athletes' diet planning, and public use. We shall provide more details on \system in the remainder of this section.

Secondly, we identify the indispensability of localized food datasets in that they take into account each country's particular cuisines in ingredients, cooking styles and varieties. Therefore, such localized food datasets could unveil valuable insights into the corresponding local human groups in dietary behaviors, which could ultimately cast light on their disease development. On account of this, we collect and curate a localized Singaporean food dataset in a systematic fashion, and the derived \dataset dataset is of high quality in terms of volume and diversity compared with its counterparts. The details of \dataset will be described in Section~\ref{sec:data collection and curation}.

Thirdly, to address both the intra-class variation issue and the inter-class resemblance issue, and mitigate the influence of hard positive/negative samples in the \dataset dataset, we introduce \scl that encourages hard positive/negative sample mining into the food recognition task and propose the \model model for recognizing food categories accurately and effectively. We shall present the concrete design of \model in Section~\ref{sec:model}.
}

\subsection{\ind{Architecture Overview}}

\begin{figure}
	\centering
	\includegraphics[width=\linewidth]{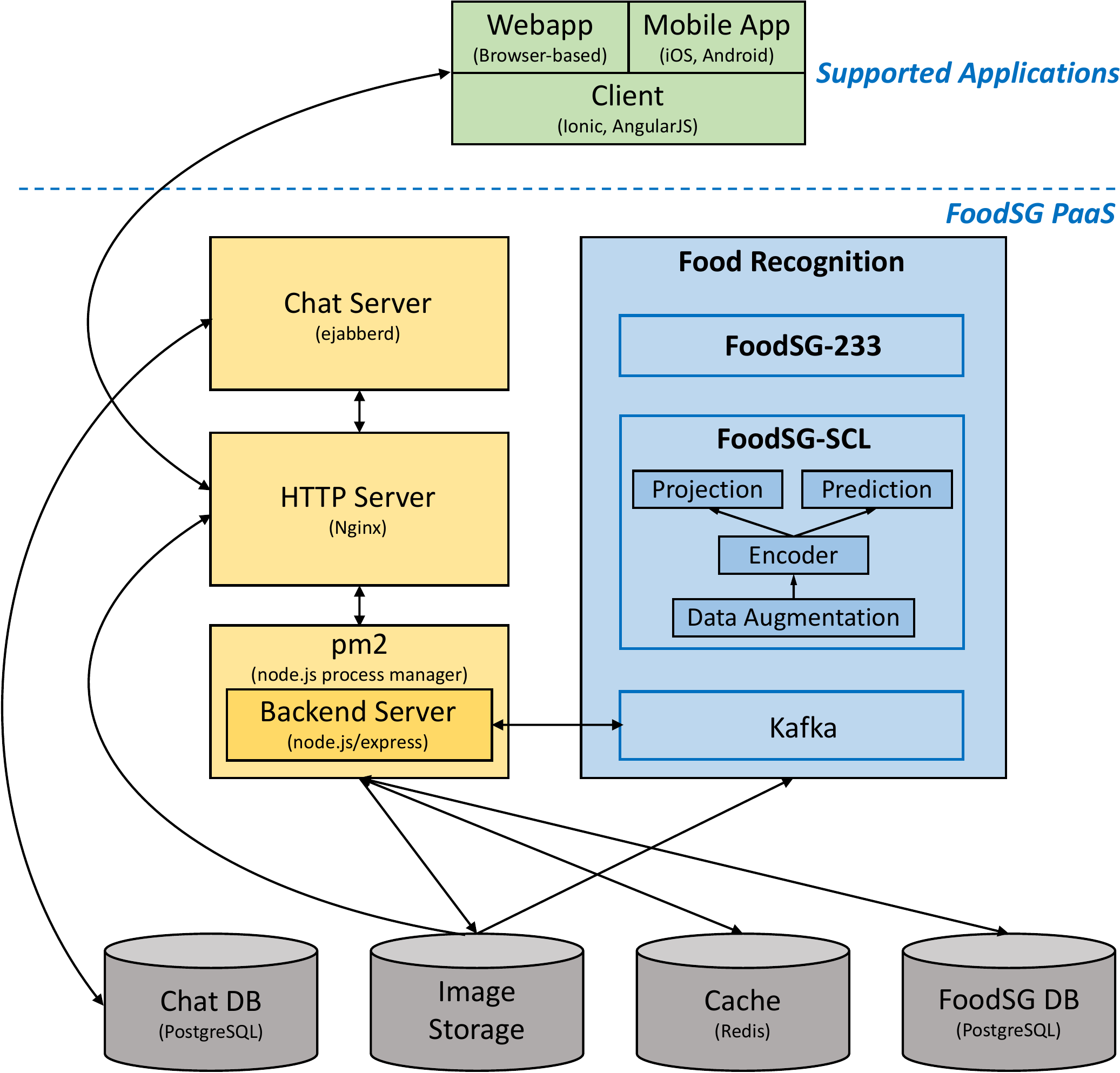}
	\caption{\ind{Overview of \system's architecture.
        }}
	\label{fig:arch}
\end{figure}

The architecture of \system's platform as a service (PaaS)  for supporting various applications is illustrated in Figure~\ref{fig:arch}.  
\system is designed for the cloud to enable adoption by local healthcare providers.
We develop its backend with node.js, and adopt Redis for caching. 
We use PostgreSQL as \system's backend database system for users' diet and exercise information collection and storage, related statistics derivation, etc. We also rely on \system's backend server to schedule \system's core component - food recognition driven by our devised \model model for analytics.
Specifically, on our curated and released \dataset dataset, we train \model integrating the Data Augmentation Module, the Encoder Module, the Projection Module, and the Prediction Module in a two-stage manner to gain the contrastive power of discriminating food images and thus, generating accurate predictions. We shall elaborate on the \dataset dataset and the \model model in Section~\ref{sec:data collection and curation} and Section~\ref{sec:model}, respectively.
Further, the data stream is managed by Apache Kafka, and we adopt ejabberd~\cite{ejabberd} to embed the instant messaging service in \system for facilitating communication between patients and their clinicians or dietitians.

Upwards, \system supports diverse applications through Angular~\cite{angular}, which provides appealing and informative user interfaces. In order to support different users' terminals covering browsers, iOS, and Android, we further introduce Ionic~\cite{ionic} for a consistent and enjoyable user experience. 

\begin{figure*}
	\begin{minipage}{\linewidth}
		\begin{minipage}{0.24\linewidth}
			\centering
			\begin{subfigure}[b]{\linewidth}
				\includegraphics[width=\linewidth]{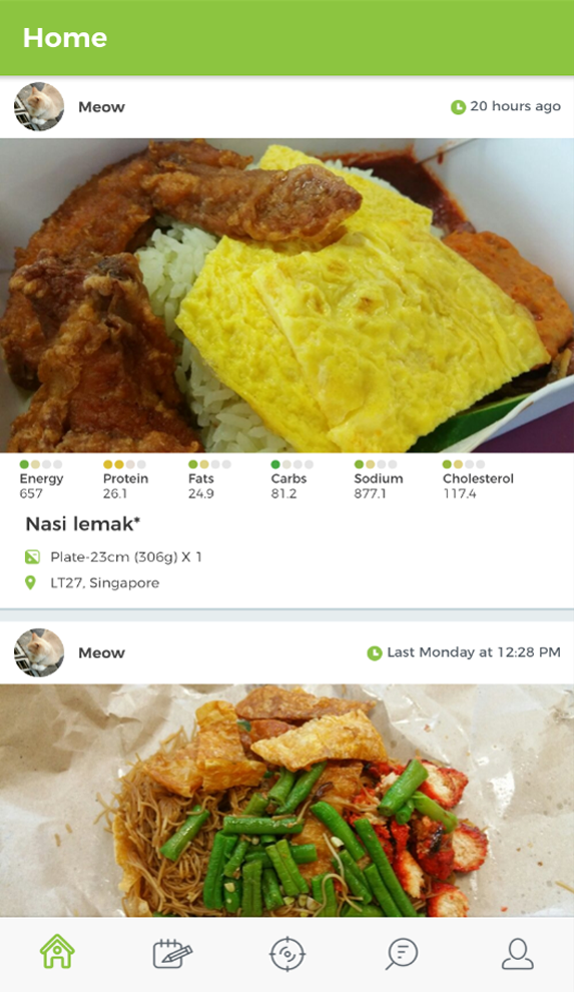}
				\caption{Food logging functionality.}
			\end{subfigure}
			\label{fig:ui1}
		\end{minipage}
		\hfill
		\begin{minipage}{0.24\linewidth}
			\centering
			\begin{subfigure}[b]{\linewidth}
				\includegraphics[width=\linewidth]{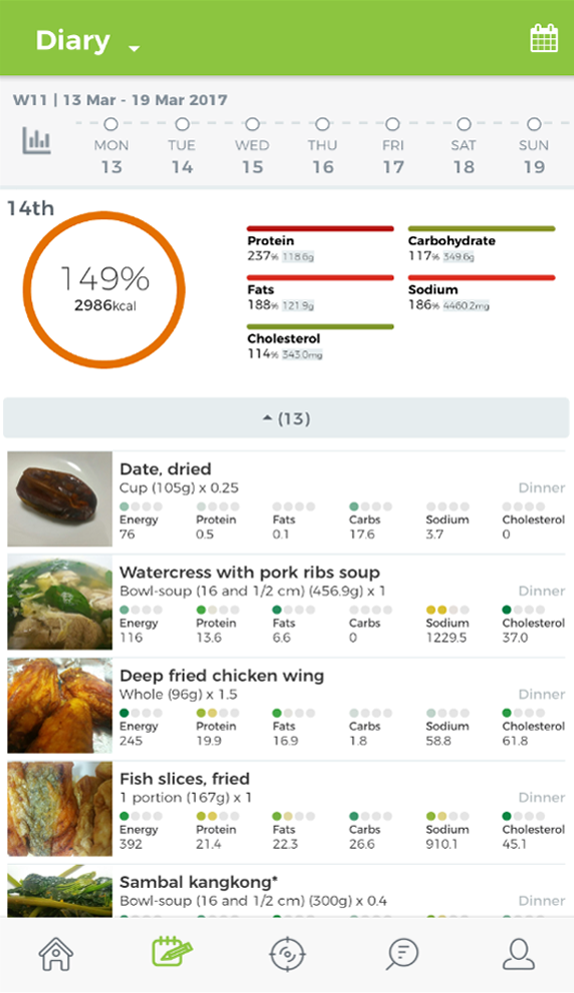}
				\caption{Diary functionality.}
			\end{subfigure}
			\label{fig:ui2}
		\end{minipage}
		\hfill
		\begin{minipage}{0.24\linewidth}
			\centering
			\begin{subfigure}[b]{\linewidth}
				\centering
				\includegraphics[width=\linewidth]{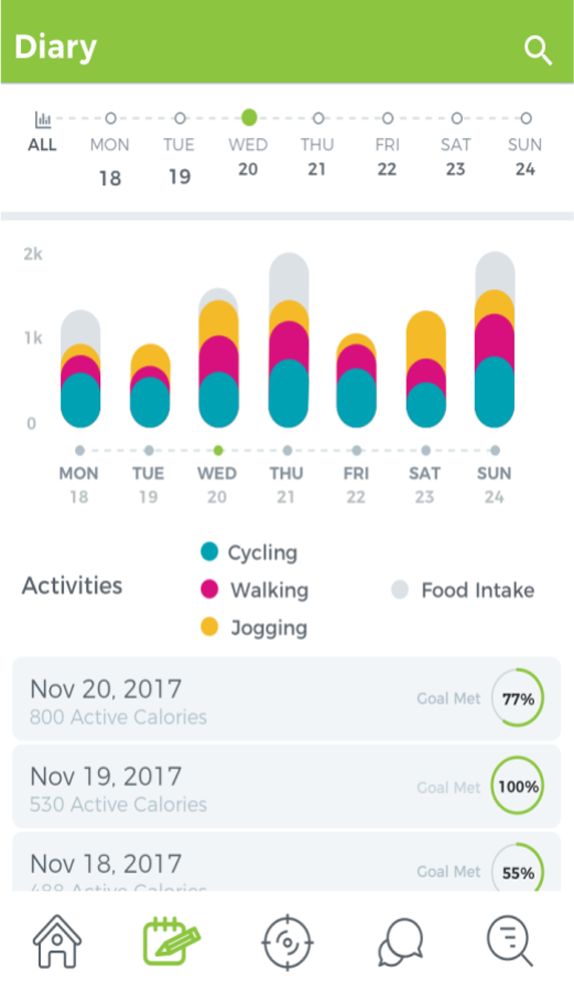}
				\caption{Activity review functionality.}
			\end{subfigure}
			\label{fig:ui3}
		\end{minipage}
		\hfill
		\begin{minipage}{0.24\linewidth}
			\centering
			\begin{subfigure}[b]{\linewidth}
				\centering
				\includegraphics[width=\linewidth]{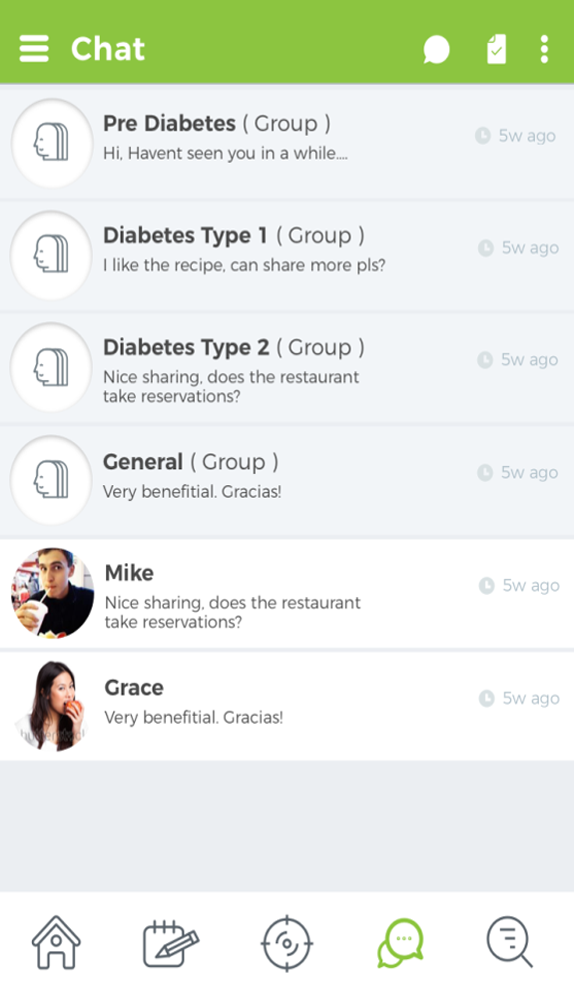}
				\caption{Chat functionality.}
			\end{subfigure}
			\label{fig:ui4}
		\end{minipage}
	\end{minipage}
	\vspace{0mm}
	\caption{\ind{Screenshots of \app powered by \system demonstrating its key functionalities.}}
	\label{fig:ui}
\end{figure*}

\ind{In Figure~\ref{fig:ui}, we present the screenshots of the ``\app'' app powered by \system for public use to demonstrate four key functionalities of \system in design below.
	
\begin{itemize}[leftmargin=*]
	
\item \textbf{Food logging.} Taking a user's snapped photo of the consumed food as input, \system automatically detects the specific food category via its food recognition model and retrieves the corresponding nutrients accurately.

\item \textbf{Diary.} \system displays the user's food journal, including the list of logged dishes for the week ordered by meal time with concrete nutrient information and a statistical summary of each day's nutrient intake. This enables the user to review his/her diet carefully. 

\item \textbf{Activity review.} \system records a user's activities over the week and assists the user in comparing the consumed food (and hence the nutrient intake) against the exercise type and amount. This functionality stimulates the user to maintain a favorable balance between his/her diet and exercise.

\item \textbf{Chat.} \system links a user with dietitians for timely and professional advice on disease prevention and monitoring, if necessary, through reviewing his/her diet diary and exercise history. \system further connects the user to other users with similar interests in a group chat for mutual encouragement.

\end{itemize}

Equipped with these functionalities, the \system platform is capable and flexible of incubating diverse, multifunction applications with varied focuses and concerns. We introduce several representative application scenarios in the next subsection.}

\subsection{\ind{Application Scenarios}}
The Singapore MOH launched a campaign to combat diabetes in April 2016, with the aim of facilitating a shared whole-of-nation endeavor to reduce the diabetes burden in Singapore~\cite{bee2022singapore, ow2021war}. Afterward, diabetes, hypertension, and hyperlipidemia are collectively considered 3H problems, which are in urgent need of management in Singapore~\cite{AI_in_Health_for_3H}.

Motivated by this, we explore and investigate food image analysis aided by dietary nutrient intake, as an infrastructure to assist in stopping or slowing the progression and complication development in these prevalent diseases.
Developed up till now, our 
established platform \system has supported a broad range of healthcare-oriented application scenarios in Singapore as a service.
In this subsection, we present four representative ongoing scenarios, 
spanning three different levels from clinical medicine to healthcare, and further to public use.

\vspace{2mm}
\noindent
\textbf{3H prevention.}
As introduced, 3H refers to (i) hyperglycemia (diabetes), (ii) hypertension (high blood pressure), and (iii) hyperlipidemia (high cholesterol)~\cite{AI_in_Health_for_3H}. If a person develops these 3H problems without proper management, his/her lifetime risk of heart disease, stroke, kidney failure, etc could be increased greatly,
leading to long-term medical attention and financial burden, and diminished quality of life. 
A practical strategy for combating 3H problems lies in advocating healthier lifestyles, e.g., to exercise more, to eat healthier food, so that we can manage or even prevent 3H.

Let's take hyperglycemia, i.e., diabetes as an example, the tackling of which is a top priority in Singapore~\cite{SG_combat_3H}.
Since
prevention is always better than cure, we aim to incentivize people to adopt healthy lifestyles. 
In the Lifestyle Intervention (LIVEN) program in the Ng Teng Fong General Hospital in Singapore, we develop the JurongHealth Food Log (\dmapp) app powered by \system to urge prediabetic patients to make behavioral lifestyle changes that are sustainable in the long term. We enroll the patients who are diagnosed with prediabetes, which is a pre-diagnosis of diabetes, indicating that the patient's blood sugar exceeds a normal range, but not reaching diabetic levels. $14.4\%$ of Singapore's population is affected by prediabetes, among whom one third will progress to type 2 diabetes in eight years eventually~\cite{JH_study}.
Fortunately, prediabetes is reversible with long-term lifestyle changes, such as disciplined diets and exercises. 

In the program, we require the patients to use \dmapp in their everyday life to record their diets and exercises.
\dmapp harnesses state-of-the-art food recognition models as the driving force, detects the food types and analyses the corresponding nutrients based on patients' uploaded photos, compiles their diets and exercise amount per day into a diary so that the patients can review their diet summaries against their weight-loss goals and exercise targets.
\dmapp also embeds a health coaching component, which connects patients with the hospital's dietitians and physiotherapists to provide valuable information, from education on nutrition and fitness to personalized feedback, and further encouragement and recommendations. In this way, they can monitor patients' progress and communicate with patients seamlessly via \dmapp, during the journey to reverse prediabetes.

We obtain promising results from a 6-month study in the LIVEN program. Almost all the patients who participate in the program experience a weight loss of between $4\%$ to $5\%$ of their initial body weight supported by \dmapp~\cite{JH_study}. As an effective and essential app for 3H prevention in clinical medicine, we plan to roll out \dmapp for a larger cohort of patients in the future.

\begin{figure*}[t]
	\centering
	\includegraphics[width=\linewidth]{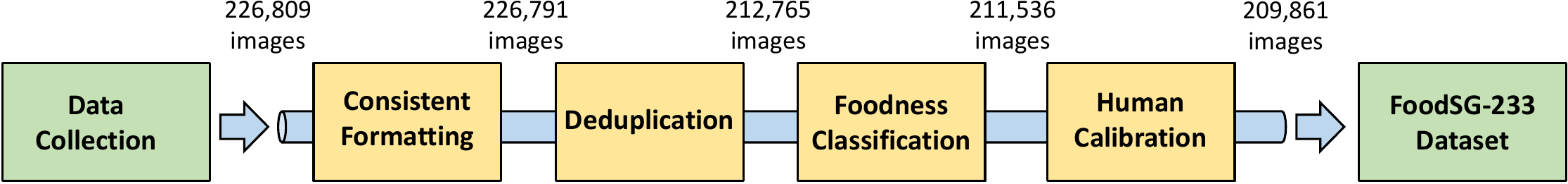}
	\caption{Data curation pipeline to derive the \dataset dataset.}
	\label{fig:data curation pipeline}
\end{figure*}

\vspace{1mm}
\noindent
\textbf{Dental care management.} 
We develop a dental care management system \dentalapp~\cite{zheng2022edental} powered by \system to record, detect and analyze users' detailed daily diets for potential risk factors for dental decay, with dentists and oral surgeons from the Faculty of Dentistry in National University of Singapore and National University Hospital. \dentalapp's key features include: (i) assisting users in logging their diets for comprehensive analysis, (ii) monitoring with attractive user interfaces and informative visual reports, (iii) triggering dietary risk alerts for dental decay when necessary, (iv) incentivizing users to continue using the system through goal setting functions, and (v) providing essential educational support to users.
Thereby, with the assistance of \dentalapp, we manage to facilitate dentists and patients to co-manage patients' daily dietary risk factors by analyzing their diet diaries. 

This scenario demonstrates the applicability of \system's service among clinical medicine, healthcare, and public use in that the end users cover dentists, their patients, and the general public who wishes to practice healthy diets for preventing dental problems. We are currently working on a user study to assess the effectiveness of \dentalapp in dental practice.

\vspace{1mm}
\noindent
\textbf{Athletes' diet planning.}
Our \system platform also provides services for Singapore Sport Institute (SSI)~\cite{SSI}
under Sport Singapore~\cite{sportsingapore},
the core purpose of which is to transform Singapore through sports and encourage individuals to live better through sports. In SSI, we have a more specific goal, i.e., we endeavor to provide the best support to athletes so that they can achieve their full potential and further fulfill their sporting aspirations. To achieve this, we turn to \system for its capability of planning the athletes' diets, guaranteeing their food diversity to achieve nutrient balance, and facilitating their exercises accordingly to maintain appropriate body weights, body mass index, etc. On the basis of the athletes' diets and exercises, we can also suggest nutrients for them to supplement.
Therefore, supported by \system, we are able to unlock the potential in the athletes and help them pursue high-performance sports. 
In this scenario, we target to deliver high-quality healthcare to athletes via \system. We are currently working on a 22-month user study to evaluate \system's positive influence on athletes' sports careers.

\vspace{1mm}
\noindent
\textbf{Public use.}
In order to serve public users, we release a public version of \system as the ``\app'' app in iOS, Android, and Web browsers, establishing a uniform and consistent user experience. We detect the food types from users' journaled entries in images or text via state-of-the-art food recognition models and calculate users' daily nutrient estimates against the standard nutritional guidelines and food composition data from the Singapore HPB~\cite{hpb}.
We also encourage users to exercise more, share food photos, and comment on them. This social feature lets the users motivate one another for maintaining a healthy lifestyle.
In this way, \app helps general users to achieve well-balanced diets, train their physical fitness, and hence, improve their health conditions and their quality of life in the long run.

In a nutshell, the aforementioned deployed scenarios showcase that our developed \system platform manages to cater to diverse applications while relying on a robust data collection and curation pipeline that will be described in the next section.

\section{\ind{Localized Singaporean Food Dataset}}
\label{sec:data collection and curation}

Localized food datasets capture the unique characteristics of each country's dish varieties, cooking styles, and food ingredients and are therefore highly indispensable and medically meaningful in contributing to people's health management in the specific country.
In this section, we elaborate on the data collection and curation process of our released localized Singaporean food dataset \dataset, which is crucial for food recognition as in Figure~\ref{fig:arch}. 
\ind{We start with the data collection and the data curation pipeline to derive the dataset and then investigate its key characteristics.}
Although our released dataset is specific to Singaporean food dishes, our proposed pipeline for data preparation is general and applicable to other types of food dishes, fostering similar research in other countries and regions.
\ind{Further, the releasing of our \dataset dataset is of vital reference value for other countries and regions, especially the ones with similar cultural food heritage as Singapore, e.g., Malaysia, in presenting a fast track for their particular food recognition app development to collect medical-grade nutrient intake information.}
\ind{Finally, we compare \dataset with existing food datasets and demonstrate its superiority in terms of scale and diversity.}

\subsection{Data Collection}

We incorporate popular ready-to-eat Singaporean food dishes by referring to HPB~\cite{hpb} for both food groups and dish names. HPB is a government organization committed to promoting healthy living in Singapore, which provides a large database of local Singaporean food and their corresponding nutrient composition~\cite{hpb_encf} measured in Per Serving Household Measure, in both grams and milliliters.
The database is hierarchically structured with different coarse-grained food groups, each of which contains fine-grained food categories.
To build a comprehensive Singaporean food dataset, we 
select 233 popular ready-to-eat Singaporean dishes from 13 main food groups and then crawl candidate images from different search engines on the Internet, including Google Images and Bing Images.
Further, in order to expand the range of images retrieved, we increase the number of crawled images using a set of synonyms for each food category in our queries. Finally, we collect $226,809$ images in total for further curation.

\subsection{Data Curation Pipeline}

In the data curation process, we follow three rules of thumb.
\begin{itemize}[leftmargin=*]
	\item Release as many images as possible so that the released dataset can support different users' practical needs.
	
	\item Add necessary preprocessing to curate and clean the dataset in order to guarantee that after cleaning, (i) the images are of high quality, and (ii) the food recognition performance is satisfactory.
	
	\item Reduce human participation as much as possible, i.e., since manual efforts are generally expensive, involve human participation only when necessary.
\end{itemize}
Driven by these principles, we design a pipeline shown in Figure~\ref{fig:data curation pipeline} to curate our collected data and derive the \dataset dataset. Next, we introduce each curation step in detail.

\subsubsection{Consistent Formatting}
Given the $226,809$ images collected,
we first convert their formats to JPEG consistently, and then remove $10$ truncated images in the process, which leads to $226,799$ images left.
Meanwhile, we constrain that each image has a minimum size of $32$ $\times$ $32$ to guarantee the quality of retrieved images, and thus remove $8$ images thereafter. After consistent formatting, the $226,791$ remaining images are fed to the next curation step.

\subsubsection{Deduplication}

We then remove exact or near duplicate images within each food category. Specifically, we calculate three different hashes for each image, namely average hashing, perceptual hashing~\cite{marr1980theory}, and difference hashing. Based on the concatenation of these three hashes, we filter out similar images and keep a single image with the largest size when there are exact or near duplicate ones.
We have $212,765$ images left after deduplication.

\subsubsection{Foodness Classification}

Due to the existence of non-food images in the dataset, we further build a separate classifier to differentiate food and non-food images for filtering out the non-food ones, i.e., a foodness classification model.
Specifically, we construct the training dataset by collecting $189,705$ non-food images from the ImageNet dataset~\cite{deng2009imagenet} as negative images, and $226,037$ food images from prevailing food datasets including VIREO Food-172~\cite{chen2016deep}, UEC-Food256~\cite{kawano2014automatic} and Food101~\cite{bossard2014food} as positive images.
We then train a DenseNet169 model~\cite{huang2017densely} for foodness classification and test the model on the current curated dataset. In the meantime, we conduct a first pass of manual checking to determine whether each image is food or not, and such manual labels are then utilized as the ground truth to evaluate the foodness model's performance. 
Through this foodness classification step, $211,536$ food images are passed to the next step of the curation pipeline, with the foodness model yielding an accuracy of $82.9\%$.

\subsubsection{Human Calibration}

To guarantee the data quality, we conduct a second pass of manual checking as human calibration to correct the images that are assigned to the wrong food categories. We end up with $209,861$ images, comprising the \dataset dataset.

\subsection{\dataset}
\label{subsec:dataset}

\begin{figure}
	\begin{minipage}{\linewidth}
		\begin{minipage}{0.49\linewidth}
			\centering
			\includegraphics[width=.9\linewidth]{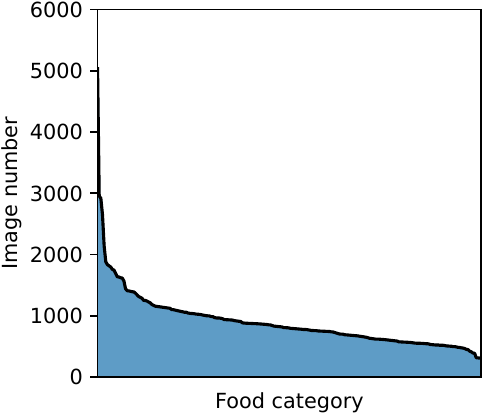}
			\caption{Sorted distribution of food image number per food category in \dataset.}
			\label{fig:categories}
		\end{minipage}
		\hfill
		\begin{minipage}{0.49\linewidth}
			\centering
			\includegraphics[width=.9\linewidth]{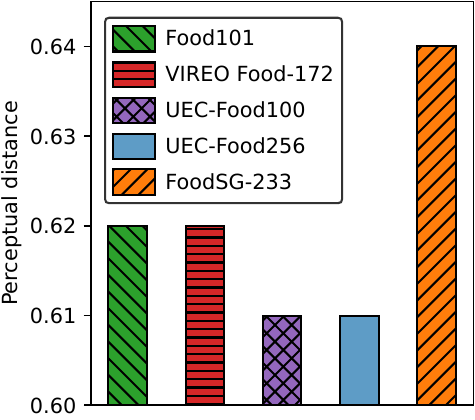}
			\caption{Diversity comparison in the perceptual distance (the larger, the more diverse).}
			\label{fig:p diversity comparison}
		\end{minipage}
	\end{minipage}
\end{figure}

\ind{After curation, \dataset contains $209,861$ images, covering $13$ food groups and $233$ food categories.
Specifically, food groups are more coarse-grained whereas food categories are more fine-grained, e.g., the group ``Sugars, sweets and confectionery'' contains categories such as ``Parfait'' and ``Popcorn''.
The detailed statistics with the category number and image number in each food group are displayed in Table~\ref{tab:dataset statistics}.
As shown, the food group ``Mixed ethnic dishes, analyzed in Singapore'' contains 129 out of 233 food categories (over $55\%$). These categories are mixed ethnic Singaporean dishes and are highly localized, such as ``laksa'' and ``lontong with sayur lodeh'' (as exemplified in Figure~\ref{fig:misclassification analysis}), which hence confirms the localization of Singapore's food consumption as well as the indispensability of localized food datasets.

We guarantee that there are at least 400 food images per food category.}
The sorted distribution of food image number per food category in \dataset is illustrated in Figure~\ref{fig:categories}, which tends to follow a power-law distribution, exhibiting the popularity of different food categories in the real world.

\begin{table}[t]
	\footnotesize
	\centering
	\caption{\dataset Data Statistics}
	\renewcommand\arraystretch{1}
	\label{tab:dataset statistics}
	\ind{\begin{tabular}{c|c|c}
			\toprule[2pt]
			Food Group & Category \# & Image \# \\ \midrule[1pt]
			Beverages & 3 & 2,128 \\ \midrule[0.1pt]
			Cereal and cereal products & 29 & 25,541 \\ \midrule[0.1pt]
			Egg and egg products & 1 & 1,000 \\ \midrule[0.1pt]
			Fast foods & 4 & 4,460 \\ \midrule[0.1pt]
			Fish and fish products & 3 & 2,041 \\ \midrule[0.1pt]
			Fruit and fruit products & 36 & 32,897 \\ \midrule[0.1pt]
			Meat and meat products & 4 & 4,136 \\ \midrule[0.1pt]
			Milk and milk products & 2 & 1,385 \\ \midrule[0.1pt]
			Mixed ethnic dishes, analyzed in Singapore & 129 & 107,986 \\ \midrule[0.1pt]
			Nuts and seeds, pulses and products & 2 & 2,822 \\ \midrule[0.1pt]
			Other mixed ethnic dishes & 4 & 3,816 \\ \midrule[0.1pt]
			Sugars, sweets and confectionery & 10 & 16,037 \\ \midrule[0.1pt]
			Vegetable and vegetable products & 6 & 5,612 \\ \midrule[0.1pt]
			Total & 233 & 209,861 \\
			\bottomrule[2pt]
		\end{tabular}}
\end{table}

\ind{Drilling down from the dataset statistics to the food images, we identify two key issues as in Figure~\ref{fig:data issues}: (i) intra-class variation of food images in terms of color, shape, and texture in the same category, and (ii) inter-class resemblance of food images among different categories. These two issues are common in real-world food datasets, which further result in the existence of hard positive/negative images. As a consequence, the performance of food recognition, which needs to mine such hard samples, is degraded.}

\subsection{Comparison with Existing Datasets}

We conduct a comparison between \dataset and several widely adopted food datasets \ind{including Food101~\cite{bossard2014food}, VIREO Food-172~\cite{chen2016deep}, UEC-Food100~\cite{matsuda2012multiple}, and UEC-Food256~\cite{kawano2014automatic}}
in Table~\ref{tab:comparison with existing datasets}.
\dataset is superior in terms of the total image number compared with its counterparts. In addition, as a localized Singaporean food dataset, \dataset has a competitively large category number and image number per category. 

\begin{table}[t]
	\setlength\tabcolsep{3pt}
	\footnotesize
	\centering
	\caption{\ind{\dataset vs. Existing Food Datasets
        }
	}
	\renewcommand\arraystretch{1}
	\label{tab:comparison with existing datasets}
	\begin{tabular}{c|c|c|c|c}
		\toprule[2pt]
		Dataset & Category \# & Image \# & Area & Image \#/Category \\ \midrule[1pt]
		Food101~\cite{bossard2014food} & 101 & 101,000 & Western & 1000 \\ \midrule[0.1pt]
		VIREO Food-172~\cite{chen2016deep} & 172 & 110,241 & Chinese & 641 \\ \midrule[0.1pt]
		UEC-Food100~\cite{matsuda2012multiple} & 100 & 14,361 & Japanese & 144 \\ \midrule[0.1pt]
		UEC-Food256~\cite{kawano2014automatic} & 256 & 25,088 & Multiple & 98 \\ \midrule[0.1pt]
		\dataset & 233 & 209,861 & Singaporean & 901 \\
		\bottomrule[2pt]
	\end{tabular}
\end{table}

\begin{figure*}
	\begin{minipage}{\linewidth}
		\begin{minipage}{0.20\linewidth}
			\centering
			\begin{subfigure}[b]{\linewidth}
				\includegraphics[width=3.7cm]{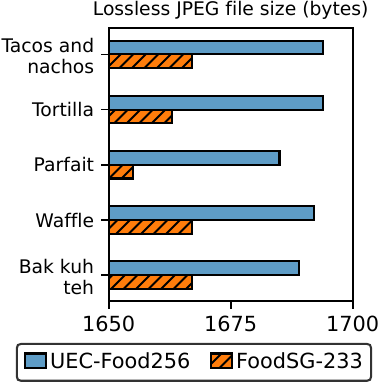}
				\caption{Diversity comparison}
			\end{subfigure}
			\label{fig:lossless comparison}
		\end{minipage}
		\hfill
		\begin{minipage}{0.39\linewidth}
			\centering
			\begin{subfigure}[b]{\linewidth}
				\centering
				\includegraphics[width=6.8cm]{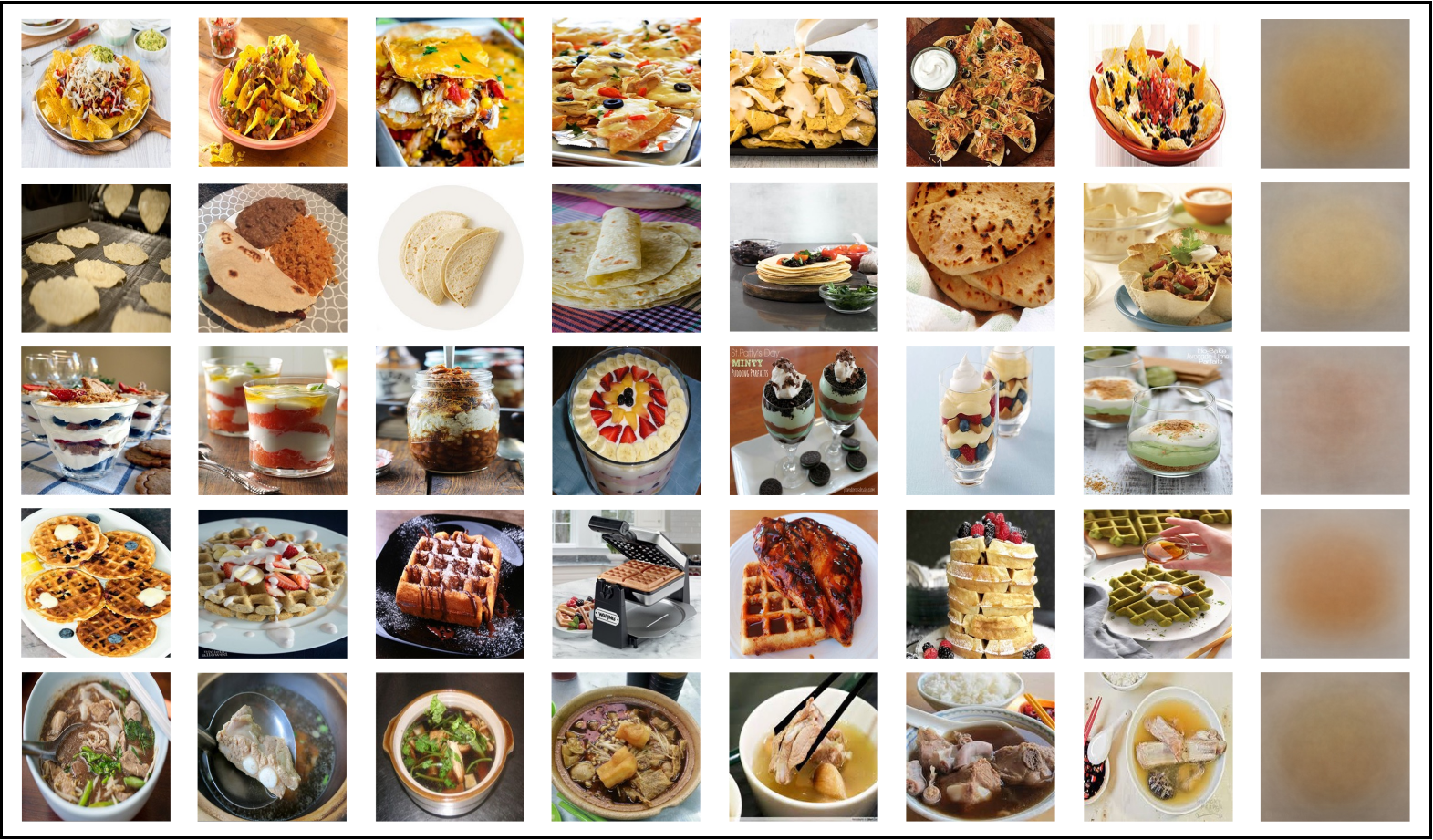}
				\caption{\dataset}
			\end{subfigure}
			\label{fig:foodlg lossless}
		\end{minipage}
		\hfill
		\begin{minipage}{0.39\linewidth}
			\centering
			\begin{subfigure}[b]{\linewidth}
				\centering
				\includegraphics[width=6.8cm]{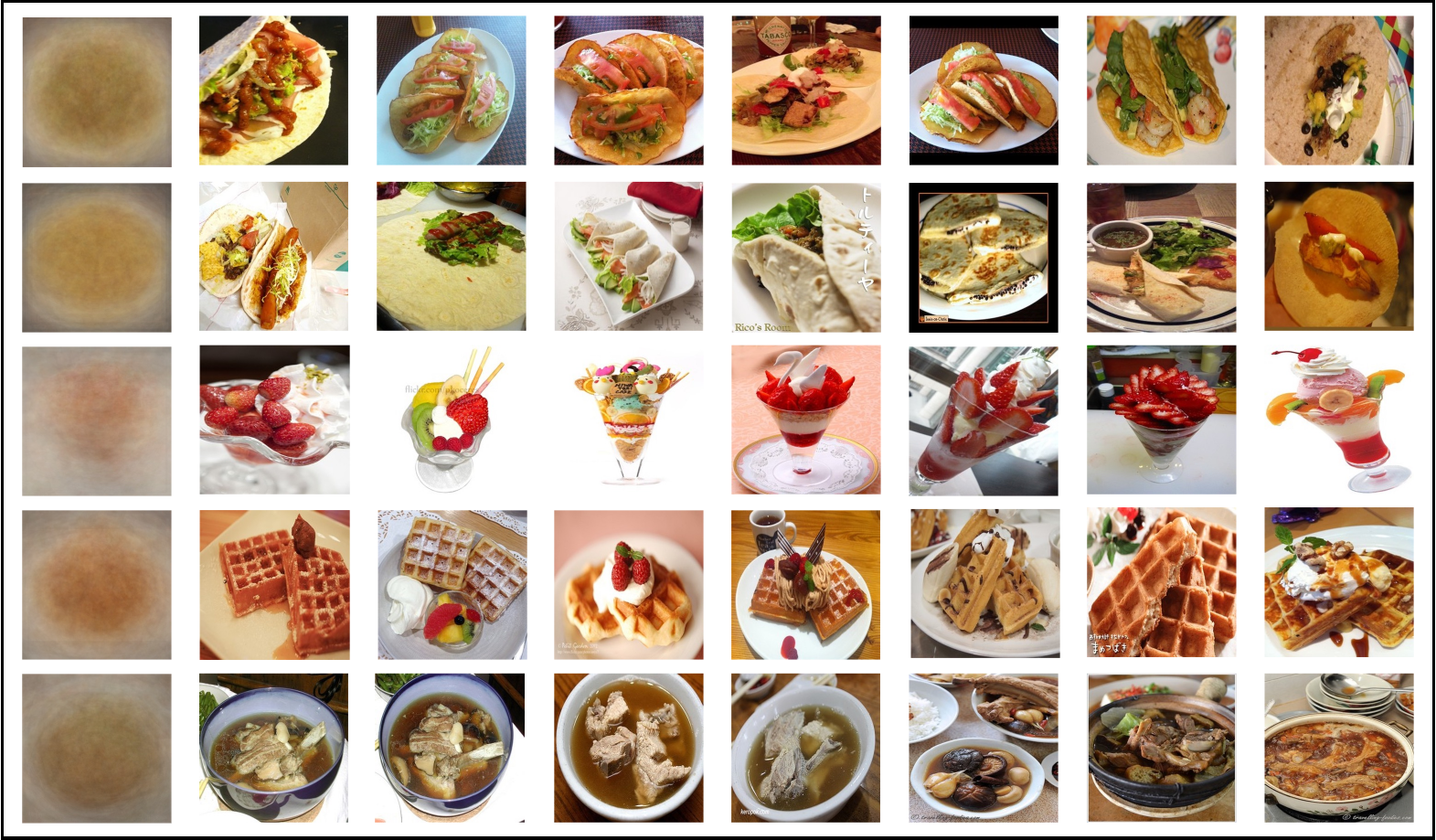}
				\caption{UEC-Food256}
			\end{subfigure}
			\label{fig:uec256 lossless}
		\end{minipage}
	\end{minipage}
	\caption{\dataset contains diversified images. (a) Diversity comparison in the lossless JPEG file sizes of five food categories' average images, i.e., the smaller the value, the more diverse. (b) Each compared food category's example images and the average image in \dataset. (c) Each compared food category's example images and the average image in UEC-Food256.}
	\vspace{2mm}
	\label{fig:jpg lossless}
\end{figure*}

As illustrated in Figure~\ref{fig:data issues}, the same food category tends to exhibit varied appearances in the real world. Hence, it is desired to release a representative dataset that covers the variations per food category and thereby, assists in building the food recognition model to tackle all sorts of user-uploaded food images effectively. With this goal, \dataset is constructed to contain food images with a high intra-class variation in terms of color, shape, and texture in the collection process. To validate the superiority of \dataset in this aspect, we conduct a quantitative diversity comparison from two perspectives below.

To start with, we employ a novel diversity measure based on a perceptual distance calculated via deep visual representations~\cite{zhang2018unreasonable}. We use the open-source model to calculate the pairwise distance within each food category, and then average different categories' distances as the diversity metric. This metric has a value range of $[0, 1]$, and the larger the value is, the more diverse the dataset is in appearance. The comparison is shown in Figure~\ref{fig:p diversity comparison}, where \dataset exhibits the highest value among all the datasets in the perceptual distance, confirming its diversity.

We take a step further to calculate the average image of each food category, upon which we can compute the lossless JPEG file size to measure the dataset diversity from another perspective. This metric reflects the information amount in an image. The underlying rationale is that a food category with more diversified images corresponds to an average image that is vaguer, whereas a category with less diversified images corresponds to a clearer average image, e.g., with a more structured shape or a sharper appearance~\cite{deng2009imagenet}. The average images are 256 $\times$ 256 and compressed into the lossless JPEG file format, \ind{and the average image file size is in bytes.}
Therefore, a more diversified food image set should exhibit a smaller size of the average image's lossless JPEG.

We compare \dataset with UEC-Food256 which has the most shared food categories. The results are illustrated in Figure~\ref{fig:jpg lossless}, including the comparison in the lossless JPEG file sizes of the average images for five food categories, the corresponding example food images, and the average images, respectively. As shown, \dataset provides more diversified food images than UEC-Food256.

\section{\ind{Effective Food Recognition}}
\label{sec:model}

In this section, we introduce our proposed \model model, which is essential to facilitate effective food recognition in \system as discussed in Section~\ref{sec:foodsg}. 
\ind{We first present its overview and then elaborate on each of its modules. Finally, we demonstrate the optimization process of \model.}

\subsection{Model Overview}

\begin{figure}
	\centering
	\includegraphics[width=0.75\linewidth]{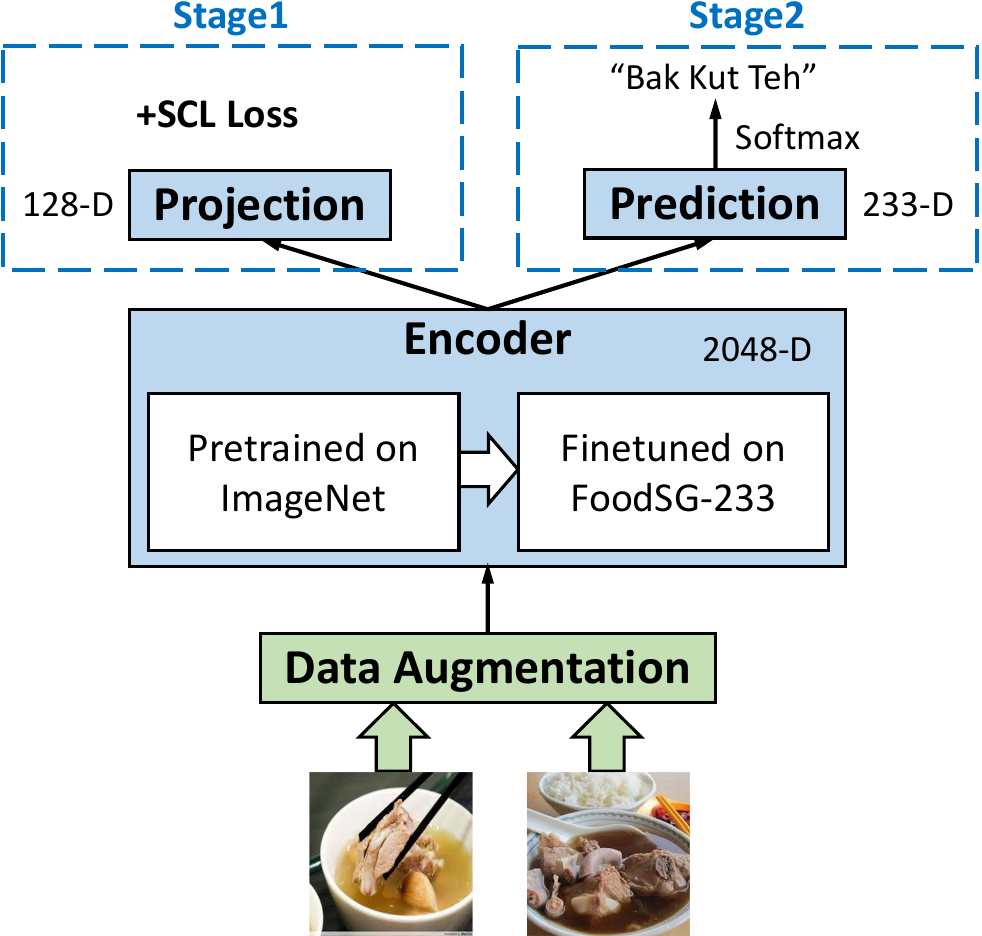}
    \vspace{-2mm}
	\caption{\model model for effective food recognition.
        }
	\label{fig:model}
\end{figure}

The overview of \model based on \scl~\cite{khosla2020supervised} for food recognition is shown in Figure~\ref{fig:model}. \model consists of four modules for data augmentation, encoding, projection, and prediction, respectively. 
The design rationale is that we first augment each data sample in the Data Augmentation Module to generate its multiple views, which facilitates \model to learn generalizable and robust representations. The augmented samples are then input to the Encoder Module to unveil the underlying information in the augmented samples, using the state-of-the-art models pretrained on the ImageNet dataset, and then finetuning them on our \dataset dataset. The Encoder Module generates a compact representation that is fed to both the Projection Module for further abstraction and integration of the \scl loss. Finally, the Prediction Module provides the final prediction out of $233$ food categories.

\subsection{Model Architecture}
\noindent
\textbf{Data Augmentation Module $\mathbf{Aug}(\cdot)$.} Given a batch of data as input, we first adopt two advanced data augmentation mechanisms to generate the ``multiviewed batch''~\cite{khosla2020supervised} composed of $2N$ samples, where $N$ is the original batch size. We denote this multiviewed batch as $\mathcal{K}=\{1, \dots, 2N\}$. As discussed above, such augmented data will be fed to the Encoder Module for further modeling.

In contrastive learning (CL)~\cite{chen2020simple}, a contrastive loss is proposed to maximize the agreement between each sample's different augmented view representations in the latent space. For instance, given an example $\mathbf{x}_i$, after the two data augmentations $\mathrm{Aug_1}(\cdot)$ and $\mathrm{Aug_2}(\cdot)$, we have $\mathbf{\tilde{x}}_i=\mathrm{Aug_1}(\mathbf{x}_i)$, and $\mathbf{\tilde{x}}_{\rho(i)}=\mathrm{Aug_2}(\mathbf{x}_i)$, where $i$ and $\rho(i)$ denote the two augmented samples from the same source sample. Hence, the index $i$ corresponds to the anchor sample, the positive sample set $\mathcal{P}=\{\rho(i)\}$ contains only one sample, and the remaining samples constitute the negative sample set.

\scl, short for supervised contrastive learning~\cite{khosla2020supervised}, extends the CL formulation above to include the samples falling into the same class (with the same $y$ label) as $i$ in its positive sample set:
\begin{equation}
	\mathcal{P}=\{p | p \in \mathcal{K} \land y_{p} = y_{i} \land p \neq i\}
\end{equation}
In comparison, \scl extends the self-supervised setting in CL to a fully-supervised setting, which effectively leverages the label information and meanwhile, contributes to mining hard positive/negative samples for boosted performance.

\vspace{2mm}
\noindent
\textbf{Encoder Module $\mathbf{Enc}(\cdot)$.} We then employ a state-of-the-art neural network model as the Encoder Module to extract the compact representations from the augmented data samples. For a higher capacity of this module, we use the model pretrained on ImageNet, and then finetune it on \dataset, grounded in the widely-acknowledged ability of convolutional neural networks in learning transferable features generalizable to diverse computer vision tasks~\cite{yosinski2014how}. After encoding, each sample $\mathbf{\tilde{x}}_i$ is converted to:
\begin{equation}
	\mathbf{e}_i=\mathrm{Enc}(\mathbf{\tilde{x}}_i)
\end{equation}
where $\mathbf{e}_i \in \mathbb{R}^{d_e}$ and $d_e$ is the dimension size of the generated representation. $d_e$ is set to $2048$ in our evaluation. We note that $\mathbf{e}_i$ is further normalized to fall in the unit hypersphere in $\mathbb{R}^{d_e}$ as suggested in~\cite{khosla2020supervised}, i.e., $\|\mathbf{e}_i\|=1$. Such normalization is considered to bring performance improvement to prediction tasks.

\vspace{2mm}
\noindent
\textbf{Projection Module $\mathbf{Pro}(\cdot)$.} Next, we project the derived $\mathbf{e}_i$ into a more compact representation in this Projection Module:
\begin{equation}
	\mathbf{s}_i=\mathrm{Pro}(\mathbf{e}_i)=\varphi_{MLP}(\mathbf{e}_i), \varphi_{MLP}:\mathbb{R}^{d_e} \longmapsto \mathbb{R}^{d_p}
\end{equation}
where a multi-layer perceptron (MLP) model $\varphi_{MLP}$ with one hidden layer is for projecting and abstracting the representation from $d_e$-dimensional to $d_p$-dimensional. We set $d_p$ to $128$ in the following experiments. Similarly, $\mathbf{s}_i$ is normalized as $\|\mathbf{s}_i\|=1$.

The \scl loss is then applied to pull positive samples close to each other and push them away from negative samples. As shown in Figure~\ref{fig:model}, the integration of \scl corresponds to \textbf{``Stage1''} in \model, which learns a strong backbone for \model in order to support the food recognition task. The learned $\mathbf{s}_i$ is discarded at the end of Stage1, while the output of the Encoder Module $\mathbf{e}_i$ will be used for prediction.

\vspace{2mm}
\noindent
\textbf{Prediction Module $\mathbf{Pre}(\cdot)$.} We input $\mathbf{e}_i$ to a linear prediction model $\psi$ for food recognition:
\begin{equation}
	\mathbf{q}_i=\mathrm{Pre}(\mathbf{e}_i)=\psi(\mathbf{e}_i), \psi: \mathbb{R}^{d_e} \longmapsto \mathbb{R}^{|\mathcal{C}|}
\end{equation}
where $\mathcal{C}$ denotes the set of food categories for prediction, $|\mathcal{C}|=233$ in the \dataset dataset. We further parameterize $\hat{\mathbf{y}}_i=\sigma(\mathbf{q}_i)$, where $\sigma(\cdot)$ is the softmax function. The derived $\hat{\mathbf{y}}_i$ is the predicted probability distribution over $|\mathcal{C}|$ classes and will be used for training \model in the optimization process. Besides, this Prediction Module corresponds to \textbf{``Stage2''} as in Figure~\ref{fig:model}, the focus of which lies in employing the well-trained \model model for food recognition in an effective manner.

\subsection{Optimization}

\noindent
\textbf{Stage1.}
The \scl loss for each sample $\mathbf{\tilde{x}}_i$ is:
\begin{equation}
	l_i^{scl} = - \frac{1}{|\mathcal{P}|} \sum_{p \in \mathcal{P}} \log \frac{\exp(sim(\mathbf{s}_i, \mathbf{s}_p)/T)}{\sum_{k \in \mathcal{K} \setminus \{i\}} \exp(sim(\mathbf{s}_i, \mathbf{s}_k)/T)}
	\label{eq:scl loss1}
\end{equation}
The contrastive prediction task is: given a sample $\mathbf{\tilde{x}}_i$, identify the positive samples in $\mathcal{P}$ out of all remaining samples $\mathcal{K}\setminus\{i\}$. In Equation~\ref{eq:scl loss1}, $sim(\cdot, \cdot)$ is the cosine similarity function, i.e., $sim(\mathbf{s}_i, \mathbf{s}_p)=\mathbf{s}_i^\top \mathbf{s}_p / (\|\mathbf{s}_i\| \|\mathbf{s}_p\|)$. As both representations are normalized, $sim(\mathbf{s}_i, \mathbf{s}_p)$ is equivalent to $\mathbf{s}_i^\top \mathbf{s}_p$. Moreover, $T$ denotes the temperature parameter.
The overall \scl loss in Stage1 sums all samples' losses:
\begin{equation}
	L^{scl} = \sum_{i \in \mathcal{K}} l_i^{scl}
\end{equation}
With this loss function in Stage1, \model (except the Prediction Module) can be effectively trained via gradient-based optimizers such as stochastic gradient descent (SGD).

\vspace{2mm}
\noindent
\textbf{Stage2.} For the food recognition task as a multi-class classification problem, we adopt the cross-entropy loss in Stage2.
\begin{equation}
	L^{ce} = - \frac{1}{|\mathcal{K}|} \sum_{i \in \mathcal{K}} \sum_{c \in \mathcal{C}} y_i^c \log(\hat{y}_i^c)
	\label{eq:ce loss}
\end{equation}
With the Encoder Module's output $\mathbf{e}_i$ and the Prediction Module, we can readily train the Predictor Module by optimizing this cross-entropy loss function.

Finally, after the optimization of both stages, \model is equipped with the contrastive power, which contributes to the learning of hard positive/negative samples and further to the prediction performance.
\ind{The effectiveness of \model will be evaluated in Section~\ref{sec:experiments}.}

\section{Experimental Evaluation}
\label{sec:experiments}

In this section, we introduce the experimental setup and analyze the experimental results.
After that, we analyze the experimental results and discuss the derived insights for practitioners on using the \system platform for their specific applications.

\subsection{Experimental Setup}
\label{subsec:exp set-up}

\ind{We evaluate the effectiveness of \model that integrates the \scl mechanism into food recognition on the \dataset dataset. Next, we introduce the evaluated models, the implementation details, and the experimental environment as follows.}

\noindent
\textbf{Evaluated Models.} We select five state-of-the-art image models for evaluation and compare their performance against integrating them into \model as the Encoder Module, in order to validate the efficacy of \scl in \model.
\begin{itemize}[leftmargin=*]
	\item \textbf{ResNet50}~\cite{he2016deep} presents a residual learning framework that is easier to train and optimize, bringing considerably improved accuracy on numerous visual recognition tasks.
	
	\item \textbf{DenseNet169}~\cite{huang2017densely} establishes direct connections between each layer and every other layer in a feed-forward manner, which enables the model to scale to a larger number of layers easily without causing difficulties in optimization. \ind{Thus, DenseNet169 serves as a feature extractor for different computer vision tasks.}
	
	\item \textbf{MobileNetV2}~\cite{sandler2018mobilenetv2} improves the efficiency of mobile models via an inverted residual structure and emphasizes the importance of linear bottlenecks to maintain the representation power.
	
	\item \textbf{EfficientNetV2}~\cite{tan2021efficientnetv2} employs optimization techniques, including training-aware neural architecture search, model scaling, and progressive learning, for accuracy, training speed, and parameter efficiency, respectively.
	
	\item \textbf{ConvNeXt}~\cite{liu2022convnet} retains the advantages of standard ConvNets in terms of simplicity and efficiency, but reexamines and redesigns the components in a ConvNet towards the modern Transformer-based models. Therefore, ConvNeXt achieves superior performance matching up to such Transformer-based models in accuracy and scalability.
\end{itemize}

\noindent
\textbf{Implementation Details.} 
The food recognition on \dataset is formulated as a $233$-class classification. We randomly split the food images per category into $80\%$, $10\%$, and $10\%$ for training, validation, and testing, respectively.
We adopt both top-1 and top-5 accuracy as evaluation metrics, and an accurate recognition model should yield high results in both metrics. We select the hyperparameters that achieve the best performance on validation data and apply such settings to the testing data for reporting the average experimental results of three independent runs.

For each of the aforementioned evaluated models, we adopt its model weights pretrained on ImageNet and then finetune them on \dataset for evaluation. We train these models via SGD~\cite{sutskever2013on} with a learning rate of $0.05$ and a weight decay of $0.0001$. We adopt batch size differently across different models, as a larger model corresponds to a smaller batch and we set the batch size as large as possible within the device memory capacity. Specifically, the batch size is $512$ in EfficientNetV2, $1024$ in DenseNet169 and ConvNeXt, $2048$ in ResNet50 and MobileNetV2, respectively.

For validating the effectiveness of \model, we integrate the evaluated models as the Encoder Module of \model, start with the same model parameters pretrained on ImageNet and then finetune on \dataset with the other essential modules shown in Figure~\ref{fig:model}.
SGD is used as the optimizer with the learning rate set to $0.1$ and the epoch number set to $200$. As \model utilizes two views for each sample, its batch size needs to be halved accordingly, \ind{i.e., $256$ for EfficientNetV2, $512$ for DenseNet169 and ConvNeXt, $1024$ for ResNet50 and MobileNetV2.}

\noindent
\textbf{Experimental Environment.} We conduct the experiments in a server with Intel(R) Xeon(R) Gold 6248R $\times$ 2, 3.0GHz, 24 cores per chip. The server has 768GB memory, and 8 NVIDIA V100 with 300GB/s NVLINK. We implement the models with PyTorch 1.12.1.

\subsection{Experimental Results}
\label{subsec:exp results}

\begin{figure}
	\centering
	\includegraphics[width=\linewidth]{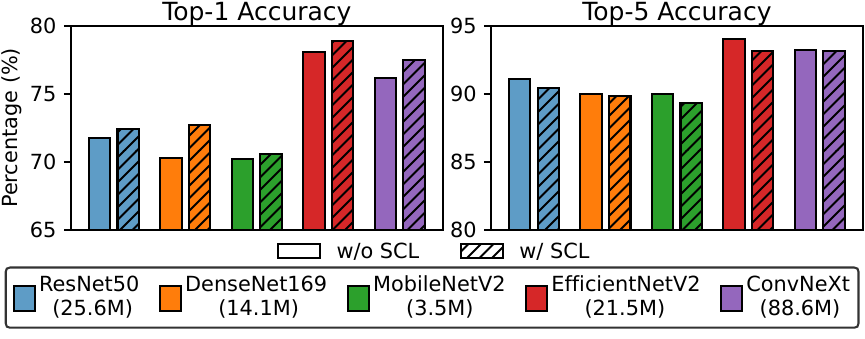}
	\caption{Comparison results of evaluated models w/o \scl and their advanced variants w/ \scl in \model. Each model's parameter number is shown in brackets.}
	\vspace{2mm}
	\label{fig:main results}
\end{figure}

In Figure~\ref{fig:main results}, we illustrate the experimental results of the evaluated image models without \scl and their respective advanced variants with \scl as well as other modules in our proposed \model.

As shown, the advanced methods with \scl outperform the ones without \scl in top-1 accuracy by a large margin consistently, which confirms the effectiveness of the \scl mechanism.
However, the integration of \scl leads to a slight degradation in top-5 accuracy. 
The underlying reasons for these phenomena will be explained in Section~\ref{subsec:exp discussions} in detail.
Among the five evaluated models, EfficientNetV2 is the most accurate in terms of both top-1 and top-5 accuracy. In addition, despite the parameter number being one magnitude smaller than the others, MobileNetV2 still achieves moderate performance in both evaluation metrics.

\begin{figure*}
	\begin{minipage}{\linewidth}
		\begin{minipage}{0.49\linewidth}
			\begin{subfigure}[b]{\linewidth}
				\begin{minipage}{0.3\linewidth}
					\centering
					\includegraphics[width=\linewidth]{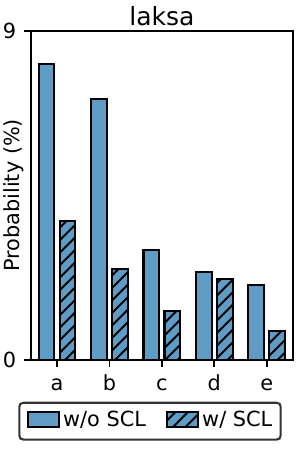}
				\end{minipage}
				\hfill
				\begin{minipage}{0.69\linewidth}
					\centering
					\includegraphics[width=\linewidth]{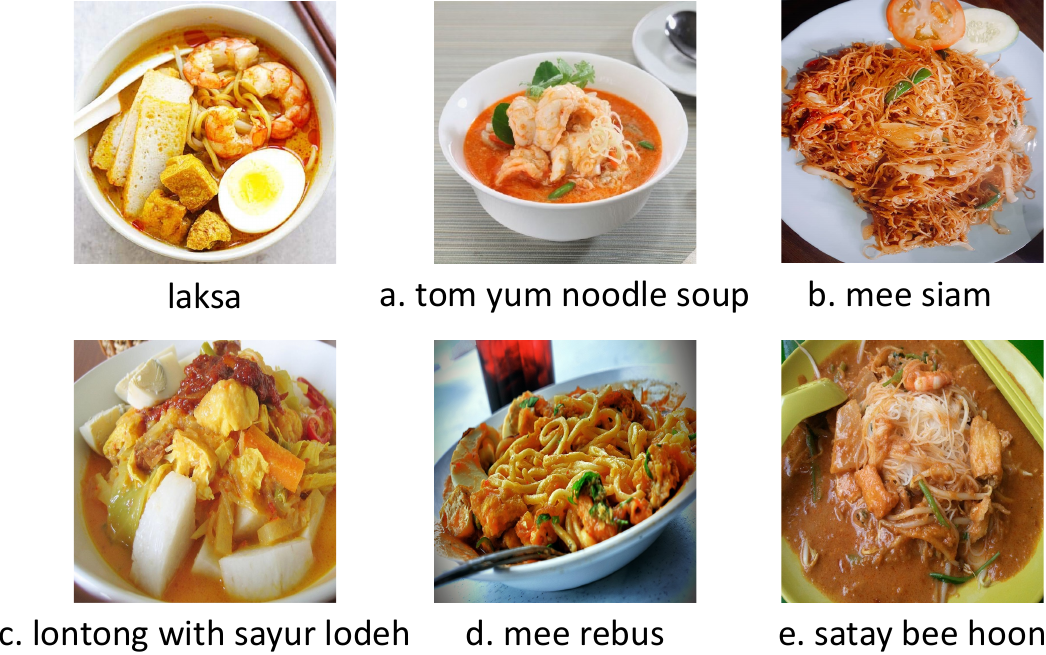}
				\end{minipage}
				\caption{Laksa}
			\end{subfigure}
		\end{minipage}
		\hfill
		\begin{minipage}{0.49\linewidth}
			\begin{subfigure}[b]{\linewidth}
				\begin{minipage}{0.3\linewidth}
					\centering
					\includegraphics[width=\linewidth]{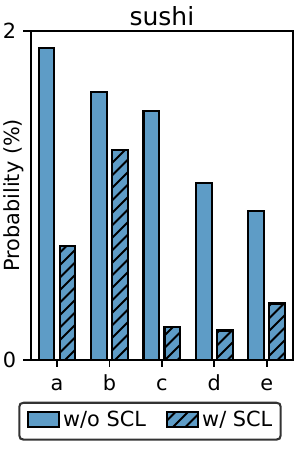}
				\end{minipage}
				\hfill
				\begin{minipage}{0.69\linewidth}
					\centering
					\includegraphics[width=\linewidth]{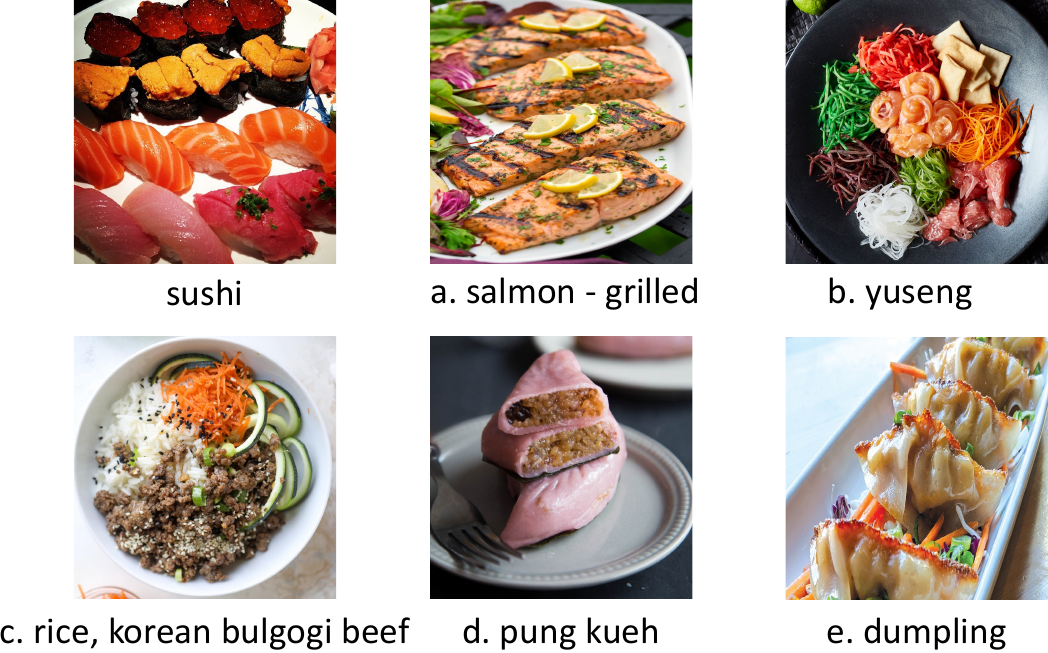}
				\end{minipage}
				\caption{Sushi}
			\end{subfigure}
		\end{minipage}
	\end{minipage}
	\vspace{0mm}
	\caption{Representative food categories and their top 5 misclassified categories.}
	\vspace{6mm}
	\label{fig:misclassification analysis}
\end{figure*}

To further investigate the intrinsic property of \scl in \model, we probe into the cases where food images are misclassified.
We choose the following two ground truth food categories to explore: (i) the representative local food category ``laksa'' which is a spicy noodle dish popular in Singapore, and (ii) ``sushi'' which is relatively more widely known.
For each ground truth category, we select the best-performing checkpoint among all evaluated models without \scl, i.e., EfficientNetV2 without \scl (according to Figure~\ref{fig:main results}), and evaluate it on all the testing samples to derive the predicted probability distribution per testing sample. We then calculate the average of all testing samples' results, based on which we illustrate the top 5 misclassified categories and their respective probabilities predicted. For comparison, we also display the results of the same five food categories derived by the most accurate model checkpoint of EfficientNetV2 with \scl.

The experimental results of both ground truth categories, including the predicted probabilities of their top 5 misclassified categories and the corresponding example food images are shown in Figure~\ref{fig:misclassification analysis}.
For both laksa and sushi, the predicted top 5 misclassified categories generated by EfficientNetV2 without \scl are highly similar to the ground truth category in terms of color, texture, and shape. 
On the contrary, EfficientNetV2 with \scl effectively reduces the predicted probabilities of these five misclassified categories in both cases, the reason for which will be explicated in Section~\ref{subsec:exp discussions}. 

\subsection{Discussions}
\label{subsec:exp discussions}

\noindent
\textbf{Top-1 accuracy vs. top-5 accuracy.}
Generally, the top 1 predicted class is used as the final result. However, in practice, the top 1 prediction may not be completely correct. To alleviate this issue, sometimes users are offered a set of five food category candidates and allowed to select the correct one with ``one more touch''. If the true food category still cannot be located within the candidate set, then the user experience will be negatively affected.
In this sense, both top-1 accuracy and top-5 accuracy are crucial in food recognition, and they measure how accurately we can recognize the food dishes in the released \dataset dataset.

As illustrated in Figure~\ref{fig:main results}, the highest top-1 accuracy is $78.87\%$ yielded by EfficientNetV2 with \scl.
Besides, EfficientNetV2 without \scl performs best in top-5 accuracy, i.e., $94.06\%$, whereas EfficientNetV2 with \scl achieves $93.13\%$ with a minor decrease. This means that in most cases, the returned five candidate food categories successfully contain the ground truth
and users could select the correct one with one more touch conveniently.
Both findings validate the effectiveness of \model for food recognition.

\vspace{1mm}
\noindent
\textbf{\scl in food recognition.}
Food recognition is challenging in that certain food categories highly resemble each other and it could be difficult to distinguish between them even for humans. Furthermore, such subtle differences in the appearance of the two categories may correspond to a large difference in their constituent nutrients, which will exert a huge impact on diet assessment. This consequently bolsters the demand for accurate food recognition when handling visually similar food images.

\scl is able to meet this demand. Specifically, \scl's underlying principle lies in drawing samples with the same label close while separating them away from samples with other labels. This in nature improves top-1 accuracy, which is shown in Figure~\ref{fig:main results} where the integration of \scl brings performance benefits consistently.

However, such benefits of \scl do not come for free.
The models without \scl adopt the standard cross-entropy loss and generate the top 5 categories that are the most similar to the ground truth category in appearance shown in Figure~\ref{fig:misclassification analysis}, due to the representation learning capability of convolutional neural networks.
Different from them, \scl-based models tend to push the negative samples with different labels away from the positive samples, in spite of their resemblance in appearance. Therefore, with \scl, the predicted probabilities of the same five similar categories will be reduced to a large extent illustrated in Figure~\ref{fig:misclassification analysis}, resulting in slightly compromised top-5 accuracy as observed in Figure~\ref{fig:main results}.

According to our real-world practice, we firmly believe that top-1 accuracy should outweigh top-5 accuracy in food recognition regardless of whether we allow users to select further from a candidate category set or not. Hence, we introduce the \scl mechanism into \model to reduce the disturbing influence exerted by similar yet incorrect categories, providing excellent performance gain in top-1 accuracy.

\vspace{2mm}
\noindent
\textbf{Edge Computing.}
In healthcare-oriented applications, there are certain scenarios where users' data cannot be processed in a centralized manner, e.g., users are not willing to upload their food photos or share their nutrient intake information for privacy concerns. This requires the computation involved in food recognition to be performed on edge devices, such as mobile phones, which is thus affected by the computing capacity and space limitation of the devices. Under such scenarios, more lightweight models are preferred as the Encoder Module of \model. For instance, MobileNetV2 exhibits a prominent advantage of parameter number being one magnitude smaller than other models, at the expense of top-1 accuracy decreasing from $78.87\%$ (by EfficientNetV2 with \scl) to $70.55\%$.

\section{Related Work}
\label{sec:related work}

\subsection{\ind{Healthcare Analytics}}

\ind{
Healthcare analytics covers numerous analytic tasks with a broad range of healthcare data spanning cost and claim data, pharmaceutical data, research and development data, clinical data, and patients' behavior and sentiment data such as patients' consumed food data. Among these types of data, clinical data that is generally collected from electronic medical records (EMR), serves as a fundamental data source to provide more effective healthcare delivery with medical insights. 
As a systematized collection of patients' electronically-stored health information in a digital format, EMR data is heterogeneous, recording patients' static characteristics such as age and gender, time-series structured information, e.g., laboratory tests, prescribed diagnoses, and unstructured features including medical imaging data and clinicians' diagnosis summary notes. With its increasing availability in recent years, EMR data becomes a crucial driving force for diverse healthcare analytic approaches, bringing promising benefits spanning more accurate diagnosis and prognosis, more optimal patient management, and possible medical research breakthrough.  

Existing studies on EMR data analytics propose different techniques to model EMR data more effectively, addressing its representative challenges meanwhile.
For instance, to take into account the temporal relationships and alleviate the lack of interpretability, RETAIN~\cite{choi2016retain} models the time-series EMR data in a reverse time order to mimic clinicians' behaviors in clinical practice and devises both the visit-level attention and the feature-level attention to facilitate accurate analytics and interpretability simultaneously. Similarly, Dipole~\cite{ma2017dipole} adopts an attention-based bidirectional recurrent neural network (RNN), which interprets the diagnosis predictions on a visit level. 
Further, several studies target to capture the irregular time intervals of EMR data, which are attributed to patients' irregular visits to the hospital. Such irregularity is addressed in T-LSTM~\cite{baytas2017patient} via a time-aware long short-term memory (LSTM) model~\cite{hochreiter1997long}, whereas mitigated by a modified gated recurrent unit (GRU) model~\cite{cho2014learning} to capture the feature-level irregular time span information in~\cite{zheng2017capturing}.
In addition, feature interactions are considered when modeling the EMR data for boosted performance and improved interpretability. Specifically, ARMOR~\cite{cai2021arm-net} captures the feature-feature interactions explicitly and interprets the prediction results via the importance of each feature-feature interaction, i.e., cross-feature. ELDA~\cite{cai2022elda} further contributes to more fine-grained interpretability with the importance of feature-feature interactions across time.
}

\subsection{\ind{Healthcare Applications}}

\ind{
In healthcare analytics, researchers propose advanced solutions to facilitate high-quality healthcare delivery in numerous applications. Among these applications, some are relatively fundamental, including (i) cohort analysis~\cite{glenn2005cohort, zhang2021grasp} to analyze the behaviors of different patient groups for testing hypotheses of new treatments or comparing patients' reactions to a certain treatment; (ii)  computational phenotyping~\cite{ho2014marble, kale2015causal} to capture the meaningful patterns in EMR data through learning latent representations in low-dimensional space in order to yield boosted performance for downstream applications; (iii) medical image analysis~\cite{balafar2010review, zhu2021towards} to develop computational approaches to address problems pertaining to medical images, such as magnetic resonance imaging data, X-ray and computed tomography scans, in order to facilitate their use in clinical care.

Grounded on these fundamental applications, healthcare analytics serves as a basis for predictive applications spanning chronic disease progression modeling~\cite{wang2014unsupervised, zheng2017resolving, zheng2017capturing, zheng2021pace}, disease diagnosis~\cite{che2015deep, lipton2015learning, zheng2020tracer, zheng2022dyhealth}, mortality prediction~\cite{che2015deep, zheng2017resolving, zheng2017capturing, zheng2020tracer, cai2022elda, zheng2022dyhealth}, and readmission prediction~\cite{caruana2015intelligent, barbieri2020benchmarking}, among others.
}

\subsection{Food Computing}

\ind{Besides the aforementioned healthcare applications, food computing is another integral application, which takes food data as input for analysis. 
Specifically,} as an interdisciplinary area, food computing targets to apply computational methods to analyze heterogeneous sources of food data for supporting diverse food-related investigations in healthcare, gastronomy, and agronomy~\cite{min2019survey}. \ind{Among the involved tasks, food recognition, which predicts the food items contained in food images, is a cornerstone of other more advanced tasks, such as nutrient intake analysis, and food recommendation, for addressing food-related issues from multiple perspectives.}

Prior studies in the early stage exploit hand-crafted features for food recognition.
\ind{For instance, in~\cite{chen2009pittsburgh}, color histograms and SIFT features are used as the image representations, while in~\cite{yang2010food}, the statistics of pairwise local features are utilized to mine the underlying spatial relationships between different ingredients in food images. Later on, the bag-of-features model is employed to describe the food images for boosted performance~\cite{anhimopoulos2014food}.}
Recently, due to the record-breaking performance achieved by deep learning models in numerous areas such as computer vision~\cite{lecun2015deep, ooi2015singa, wang2016database}, various convolutional neural network models~\cite{kagaya2014food, wu2016learning} are employed in food recognition for extracting the features in food images and hence, delivering more satisfactory recognition performance than the aforementioned traditional approaches.

In the meantime, a large number of food-related datasets are released to further promote the development of food computing. For example, Food101~\cite{bossard2014food} is constructed to cover $101$ food categories, containing $101,000$ images in total, and most of its images correspond to western food. Different from Food101, VIREO Food-172~\cite{chen2016deep} turns its focus to Chinese food and includes $172$ food categories, and $110,241$ images, respectively. Moreover, a Japanese food dataset UEC-Food100~\cite{matsuda2012multiple} is built to incorporate 100 popular Japanese food categories. This UEC-Food100 dataset is then expanded to UEC-Food256~\cite{kawano2014automatic} introducing more food categories from other countries, namely French, Italian, American, Chinese, Thai, Vietnamese, and Indonesian.
These released multifarious food-related datasets present researchers with ample opportunity for food computing and analysis. However, these datasets have their specific geographic areas and therefore, only include particular types of cuisines.
Different from existing work, driven by \system, we investigate Singaporean food dishes and release the localized dataset \dataset, which is of vital significance to \system's provided healthcare services for Singaporeans. 

\section{Conclusions}
\label{sec:conclusions}

\ind{In order to surmount the difficulty of retrieving medical-grade nutrient intake of Singaporeans for advocating healthier lifestyles, we abstract the shared requirements among diverse applications and develop the dietary nutrient-aided healthcare platform \system as a service to support all sorts of application scenarios in Singapore.
We collect, curate, and release a Singaporean food dataset \dataset systematically to promote future research directions for the data management community in food computing.
Through delving into \dataset to analyze its issues of intra-class dissimilarity and inter-class similarity, we propose to integrate \scl into food recognition and devise the \model model accordingly to facilitate the learning from hard positive/negative samples for performance gains.
By evaluating \model from multiple perspectives, we deliver fresh insights and valuable experience in healthcare applications to practitioners.}

\newpage

\bibliographystyle{ACM-Reference-Format}
\bibliography{main}


\begin{thebibliography}{70}


\ifx \showCODEN    \undefined \def \showCODEN     #1{\unskip}     \fi
\ifx \showDOI      \undefined \def \showDOI       #1{#1}\fi
\ifx \showISBNx    \undefined \def \showISBNx     #1{\unskip}     \fi
\ifx \showISBNxiii \undefined \def \showISBNxiii  #1{\unskip}     \fi
\ifx \showISSN     \undefined \def \showISSN      #1{\unskip}     \fi
\ifx \showLCCN     \undefined \def \showLCCN      #1{\unskip}     \fi
\ifx \shownote     \undefined \def \shownote      #1{#1}          \fi
\ifx \showarticletitle \undefined \def \showarticletitle #1{#1}   \fi
\ifx \showURL      \undefined \def \showURL       {\relax}        \fi
\providecommand\bibfield[2]{#2}
\providecommand\bibinfo[2]{#2}
\providecommand\natexlab[1]{#1}
\providecommand\showeprint[2][]{arXiv:#2}

\bibitem[\protect\citeauthoryear{??}{JH_}{2019}]%
        {JH_study}
 \bibinfo{year}{2019}\natexlab{}.
\newblock
  \bibinfo{howpublished}{\url{https://www.todayonline.com/singapore/get-snap-happy-new-app-pre-diabetics-help-manage-health-condition}}.
\newblock


\bibitem[\protect\citeauthoryear{??}{SG_}{2021}]%
        {SG_combat_3H}
 \bibinfo{year}{2021}\natexlab{}.
\newblock
  \bibinfo{howpublished}{\url{https://www.pmo.gov.sg/Newsroom/PM-Lee-Hsien-Loong-WHO-Global-Diabetes-Compact}}.
\newblock


\bibitem[\protect\citeauthoryear{??}{AI_}{2022}]%
        {AI_in_Health_for_3H}
 \bibinfo{year}{2022}\natexlab{}.
\newblock
  \bibinfo{howpublished}{\url{https://aisingapore.org/grand-challenges/}}.
\newblock


\bibitem[\protect\citeauthoryear{??}{haw}{2022}]%
        {hawkercentre}
 \bibinfo{year}{2022}\natexlab{}.
\newblock
  \bibinfo{howpublished}{\url{https://en.wikipedia.org/wiki/Hawker_centre}}.
\newblock


\bibitem[\protect\citeauthoryear{??}{SSI}{2022}]%
        {SSI}
 \bibinfo{year}{2022}\natexlab{}.
\newblock
  \bibinfo{howpublished}{\url{https://www.sportsingapore.gov.sg/athletes-coaches/singapore-sport-institute}}.
\newblock


\bibitem[\protect\citeauthoryear{??}{spo}{2022}]%
        {sportsingapore}
 \bibinfo{year}{2022}\natexlab{}.
\newblock \bibinfo{howpublished}{\url{https://www.sportsingapore.gov.sg/}}.
\newblock


\bibitem[\protect\citeauthoryear{??}{ang}{2022}]%
        {angular}
 \bibinfo{year}{2022}\natexlab{}.
\newblock \bibinfo{title}{Angular}.
\newblock \bibinfo{howpublished}{\url{https://angular.io/}}.
\newblock


\bibitem[\protect\citeauthoryear{??}{eja}{2022}]%
        {ejabberd}
 \bibinfo{year}{2022}\natexlab{}.
\newblock \bibinfo{title}{ejabberd}.
\newblock \bibinfo{howpublished}{\url{https://www.ejabberd.im/}}.
\newblock


\bibitem[\protect\citeauthoryear{??}{hpb}{2022a}]%
        {hpb_encf}
 \bibinfo{year}{2022}\natexlab{a}.
\newblock \bibinfo{title}{Energy \& Nutrient Composition of Food}.
\newblock
  \bibinfo{howpublished}{\url{https://focos.hpb.gov.sg/eservices/ENCF/}}.
\newblock


\bibitem[\protect\citeauthoryear{??}{hpb}{2022b}]%
        {hpb}
 \bibinfo{year}{2022}\natexlab{b}.
\newblock \bibinfo{title}{Health Promotion Board}.
\newblock \bibinfo{howpublished}{\url{https://hpb.gov.sg/}}.
\newblock


\bibitem[\protect\citeauthoryear{??}{ion}{2022}]%
        {ionic}
 \bibinfo{year}{2022}\natexlab{}.
\newblock \bibinfo{title}{Ionic}.
\newblock \bibinfo{howpublished}{\url{https://ionicframework.com/}}.
\newblock


\bibitem[\protect\citeauthoryear{??}{sgf}{2023}]%
        {sgfood}
 \bibinfo{year}{2023}\natexlab{}.
\newblock
  \bibinfo{howpublished}{\url{https://en.wikipedia.org/wiki/Singaporean_cuisine}}.
\newblock


\bibitem[\protect\citeauthoryear{??}{hea}{2023}]%
        {healthiersg}
 \bibinfo{year}{2023}\natexlab{}.
\newblock \bibinfo{howpublished}{\url{https://www.healthiersg.gov.sg/}}.
\newblock


\bibitem[\protect\citeauthoryear{Achananuparp, Lim, and Abhishek}{Achananuparp
  et~al\mbox{.}}{2018}]%
        {achananuparp2018does}
\bibfield{author}{\bibinfo{person}{Palakorn Achananuparp},
  \bibinfo{person}{Ee{-}Peng Lim}, {and} \bibinfo{person}{Vibhanshu Abhishek}.}
  \bibinfo{year}{2018}\natexlab{}.
\newblock \showarticletitle{Does Journaling Encourage Healthier Choices?:
  Analyzing Healthy Eating Behaviors of Food Journalers}. In
  \bibinfo{booktitle}{\emph{{DH}}}. \bibinfo{publisher}{{ACM}},
  \bibinfo{pages}{35--44}.
\newblock


\bibitem[\protect\citeauthoryear{Anthimopoulos, Gianola, Scarnato, Diem, and
  Mougiakakou}{Anthimopoulos et~al\mbox{.}}{2014}]%
        {anhimopoulos2014food}
\bibfield{author}{\bibinfo{person}{Marios Anthimopoulos},
  \bibinfo{person}{Lauro Gianola}, \bibinfo{person}{Luca Scarnato},
  \bibinfo{person}{Peter Diem}, {and} \bibinfo{person}{Stavroula~G.
  Mougiakakou}.} \bibinfo{year}{2014}\natexlab{}.
\newblock \showarticletitle{A Food Recognition System for Diabetic Patients
  Based on an Optimized Bag-of-Features Model}.
\newblock \bibinfo{journal}{\emph{{IEEE} J. Biomed. Health Informatics}}
  \bibinfo{volume}{18}, \bibinfo{number}{4} (\bibinfo{year}{2014}),
  \bibinfo{pages}{1261--1271}.
\newblock


\bibitem[\protect\citeauthoryear{Balafar, Ramli, Saripan, and Mashohor}{Balafar
  et~al\mbox{.}}{2010}]%
        {balafar2010review}
\bibfield{author}{\bibinfo{person}{M.~A. Balafar},
  \bibinfo{person}{Abdul~Rahman Ramli}, \bibinfo{person}{M.~Iqbal Saripan},
  {and} \bibinfo{person}{Syamsiah Mashohor}.} \bibinfo{year}{2010}\natexlab{}.
\newblock \showarticletitle{Review of brain {MRI} image segmentation methods}.
\newblock \bibinfo{journal}{\emph{Artif. Intell. Rev.}} \bibinfo{volume}{33},
  \bibinfo{number}{3} (\bibinfo{year}{2010}), \bibinfo{pages}{261--274}.
\newblock


\bibitem[\protect\citeauthoryear{Barbieri, Kemp, Perez-Concha, Kotwal,
  Gallagher, Ritchie, and Jorm}{Barbieri et~al\mbox{.}}{2020}]%
        {barbieri2020benchmarking}
\bibfield{author}{\bibinfo{person}{Sebastiano Barbieri}, \bibinfo{person}{James
  Kemp}, \bibinfo{person}{Oscar Perez-Concha}, \bibinfo{person}{Sradha Kotwal},
  \bibinfo{person}{Martin Gallagher}, \bibinfo{person}{Angus Ritchie}, {and}
  \bibinfo{person}{Louisa Jorm}.} \bibinfo{year}{2020}\natexlab{}.
\newblock \showarticletitle{Benchmarking deep learning architectures for
  predicting readmission to the ICU and describing patients-at-risk}.
\newblock \bibinfo{journal}{\emph{Scientific reports}} \bibinfo{volume}{10},
  \bibinfo{number}{1} (\bibinfo{year}{2020}), \bibinfo{pages}{1111}.
\newblock


\bibitem[\protect\citeauthoryear{Baytas, Xiao, Zhang, Wang, Jain, and
  Zhou}{Baytas et~al\mbox{.}}{2017}]%
        {baytas2017patient}
\bibfield{author}{\bibinfo{person}{Inci~M. Baytas}, \bibinfo{person}{Cao Xiao},
  \bibinfo{person}{Xi Zhang}, \bibinfo{person}{Fei Wang},
  \bibinfo{person}{Anil~K. Jain}, {and} \bibinfo{person}{Jiayu Zhou}.}
  \bibinfo{year}{2017}\natexlab{}.
\newblock \showarticletitle{Patient Subtyping via Time-Aware {LSTM} Networks}.
  In \bibinfo{booktitle}{\emph{{KDD}}}. \bibinfo{publisher}{{ACM}},
  \bibinfo{pages}{65--74}.
\newblock


\bibitem[\protect\citeauthoryear{Bee, Tai, and Wong}{Bee et~al\mbox{.}}{2022}]%
        {bee2022singapore}
\bibfield{author}{\bibinfo{person}{Yong~Mong Bee}, \bibinfo{person}{E~Shyong
  Tai}, {and} \bibinfo{person}{Tien~Y Wong}.} \bibinfo{year}{2022}\natexlab{}.
\newblock \showarticletitle{Singapore's “War on Diabetes”}.
\newblock \bibinfo{journal}{\emph{The Lancet Diabetes \& Endocrinology}}
  \bibinfo{volume}{10}, \bibinfo{number}{6} (\bibinfo{year}{2022}),
  \bibinfo{pages}{391--392}.
\newblock


\bibitem[\protect\citeauthoryear{Bossard, Guillaumin, and Gool}{Bossard
  et~al\mbox{.}}{2014}]%
        {bossard2014food}
\bibfield{author}{\bibinfo{person}{Lukas Bossard}, \bibinfo{person}{Matthieu
  Guillaumin}, {and} \bibinfo{person}{Luc~Van Gool}.}
  \bibinfo{year}{2014}\natexlab{}.
\newblock \showarticletitle{Food-101 - Mining Discriminative Components with
  Random Forests}. In \bibinfo{booktitle}{\emph{{ECCV} {(6)}}}
  \emph{(\bibinfo{series}{Lecture Notes in Computer Science},
  Vol.~\bibinfo{volume}{8694})}. \bibinfo{publisher}{Springer},
  \bibinfo{pages}{446--461}.
\newblock


\bibitem[\protect\citeauthoryear{Cai, Zheng, Ooi, Wang, and Yao}{Cai
  et~al\mbox{.}}{2022}]%
        {cai2022elda}
\bibfield{author}{\bibinfo{person}{Qingpeng Cai}, \bibinfo{person}{Kaiping
  Zheng}, \bibinfo{person}{Beng~Chin Ooi}, \bibinfo{person}{Wei Wang}, {and}
  \bibinfo{person}{Chang Yao}.} \bibinfo{year}{2022}\natexlab{}.
\newblock \showarticletitle{{ELDA:} Learning Explicit Dual-Interactions for
  Healthcare Analytics}. In \bibinfo{booktitle}{\emph{{ICDE}}}.
  \bibinfo{publisher}{{IEEE}}, \bibinfo{pages}{393--406}.
\newblock


\bibitem[\protect\citeauthoryear{Cai, Zheng, Chen, Jagadish, Ooi, and
  Zhang}{Cai et~al\mbox{.}}{2021}]%
        {cai2021arm-net}
\bibfield{author}{\bibinfo{person}{Shaofeng Cai}, \bibinfo{person}{Kaiping
  Zheng}, \bibinfo{person}{Gang Chen}, \bibinfo{person}{H.~V. Jagadish},
  \bibinfo{person}{Beng~Chin Ooi}, {and} \bibinfo{person}{Meihui Zhang}.}
  \bibinfo{year}{2021}\natexlab{}.
\newblock \showarticletitle{ARM-Net: Adaptive Relation Modeling Network for
  Structured Data}. In \bibinfo{booktitle}{\emph{{SIGMOD} Conference}}.
  \bibinfo{publisher}{{ACM}}, \bibinfo{pages}{207--220}.
\newblock


\bibitem[\protect\citeauthoryear{Caruana, Lou, Gehrke, Koch, Sturm, and
  Elhadad}{Caruana et~al\mbox{.}}{2015}]%
        {caruana2015intelligent}
\bibfield{author}{\bibinfo{person}{Rich Caruana}, \bibinfo{person}{Yin Lou},
  \bibinfo{person}{Johannes Gehrke}, \bibinfo{person}{Paul Koch},
  \bibinfo{person}{Marc Sturm}, {and} \bibinfo{person}{Noemie Elhadad}.}
  \bibinfo{year}{2015}\natexlab{}.
\newblock \showarticletitle{Intelligible Models for HealthCare: Predicting
  Pneumonia Risk and Hospital 30-day Readmission}. In
  \bibinfo{booktitle}{\emph{{KDD}}}. \bibinfo{publisher}{{ACM}},
  \bibinfo{pages}{1721--1730}.
\newblock


\bibitem[\protect\citeauthoryear{Che, Kale, Li, Bahadori, and Liu}{Che
  et~al\mbox{.}}{2015}]%
        {che2015deep}
\bibfield{author}{\bibinfo{person}{Zhengping Che}, \bibinfo{person}{David~C.
  Kale}, \bibinfo{person}{Wenzhe Li}, \bibinfo{person}{Mohammad~Taha Bahadori},
  {and} \bibinfo{person}{Yan Liu}.} \bibinfo{year}{2015}\natexlab{}.
\newblock \showarticletitle{Deep Computational Phenotyping}. In
  \bibinfo{booktitle}{\emph{{KDD}}}. \bibinfo{publisher}{{ACM}},
  \bibinfo{pages}{507--516}.
\newblock


\bibitem[\protect\citeauthoryear{Chen and Ngo}{Chen and Ngo}{2016}]%
        {chen2016deep}
\bibfield{author}{\bibinfo{person}{Jingjing Chen} {and}
  \bibinfo{person}{Chong{-}Wah Ngo}.} \bibinfo{year}{2016}\natexlab{}.
\newblock \showarticletitle{Deep-based Ingredient Recognition for Cooking
  Recipe Retrieval}. In \bibinfo{booktitle}{\emph{{ACM} Multimedia}}.
  \bibinfo{publisher}{{ACM}}, \bibinfo{pages}{32--41}.
\newblock


\bibitem[\protect\citeauthoryear{Chen, Dhingra, Wu, Yang, Sukthankar, and
  Yang}{Chen et~al\mbox{.}}{2009}]%
        {chen2009pittsburgh}
\bibfield{author}{\bibinfo{person}{Mei Chen}, \bibinfo{person}{Kapil Dhingra},
  \bibinfo{person}{Wen Wu}, \bibinfo{person}{Lei Yang}, \bibinfo{person}{Rahul
  Sukthankar}, {and} \bibinfo{person}{Jie Yang}.}
  \bibinfo{year}{2009}\natexlab{}.
\newblock \showarticletitle{{PFID:} Pittsburgh fast-food image dataset}. In
  \bibinfo{booktitle}{\emph{{ICIP}}}. \bibinfo{publisher}{{IEEE}},
  \bibinfo{pages}{289--292}.
\newblock


\bibitem[\protect\citeauthoryear{Chen, Kornblith, Norouzi, and Hinton}{Chen
  et~al\mbox{.}}{2020}]%
        {chen2020simple}
\bibfield{author}{\bibinfo{person}{Ting Chen}, \bibinfo{person}{Simon
  Kornblith}, \bibinfo{person}{Mohammad Norouzi}, {and}
  \bibinfo{person}{Geoffrey~E. Hinton}.} \bibinfo{year}{2020}\natexlab{}.
\newblock \showarticletitle{A Simple Framework for Contrastive Learning of
  Visual Representations}. In \bibinfo{booktitle}{\emph{{ICML}}}
  \emph{(\bibinfo{series}{Proceedings of Machine Learning Research},
  Vol.~\bibinfo{volume}{119})}. \bibinfo{publisher}{{PMLR}},
  \bibinfo{pages}{1597--1607}.
\newblock


\bibitem[\protect\citeauthoryear{Cho, van Merrienboer, G{\"{u}}l{\c{c}}ehre,
  Bahdanau, Bougares, Schwenk, and Bengio}{Cho et~al\mbox{.}}{2014}]%
        {cho2014learning}
\bibfield{author}{\bibinfo{person}{Kyunghyun Cho}, \bibinfo{person}{Bart van
  Merrienboer}, \bibinfo{person}{{\c{C}}aglar G{\"{u}}l{\c{c}}ehre},
  \bibinfo{person}{Dzmitry Bahdanau}, \bibinfo{person}{Fethi Bougares},
  \bibinfo{person}{Holger Schwenk}, {and} \bibinfo{person}{Yoshua Bengio}.}
  \bibinfo{year}{2014}\natexlab{}.
\newblock \showarticletitle{Learning Phrase Representations using {RNN}
  Encoder-Decoder for Statistical Machine Translation}. In
  \bibinfo{booktitle}{\emph{{EMNLP}}}. \bibinfo{publisher}{{ACL}},
  \bibinfo{pages}{1724--1734}.
\newblock


\bibitem[\protect\citeauthoryear{Choi, Bahadori, Sun, Kulas, Schuetz, and
  Stewart}{Choi et~al\mbox{.}}{2016}]%
        {choi2016retain}
\bibfield{author}{\bibinfo{person}{Edward Choi}, \bibinfo{person}{Mohammad~Taha
  Bahadori}, \bibinfo{person}{Jimeng Sun}, \bibinfo{person}{Joshua Kulas},
  \bibinfo{person}{Andy Schuetz}, {and} \bibinfo{person}{Walter~F. Stewart}.}
  \bibinfo{year}{2016}\natexlab{}.
\newblock \showarticletitle{{RETAIN:} An Interpretable Predictive Model for
  Healthcare using Reverse Time Attention Mechanism}. In
  \bibinfo{booktitle}{\emph{{NIPS}}}. \bibinfo{pages}{3504--3512}.
\newblock


\bibitem[\protect\citeauthoryear{Deng, Dong, Socher, Li, Li, and Fei-Fei}{Deng
  et~al\mbox{.}}{2009}]%
        {deng2009imagenet}
\bibfield{author}{\bibinfo{person}{J. Deng}, \bibinfo{person}{W. Dong},
  \bibinfo{person}{R. Socher}, \bibinfo{person}{L.-J. Li}, \bibinfo{person}{K.
  Li}, {and} \bibinfo{person}{L. Fei-Fei}.} \bibinfo{year}{2009}\natexlab{}.
\newblock \showarticletitle{{ImageNet: A Large-Scale Hierarchical Image
  Database}}. In \bibinfo{booktitle}{\emph{CVPR09}}.
\newblock


\bibitem[\protect\citeauthoryear{Glenn}{Glenn}{2005}]%
        {glenn2005cohort}
\bibfield{author}{\bibinfo{person}{Norval~D Glenn}.}
  \bibinfo{year}{2005}\natexlab{}.
\newblock \bibinfo{booktitle}{\emph{Cohort analysis}}.
  Vol.~\bibinfo{volume}{5}.
\newblock \bibinfo{publisher}{Sage}.
\newblock


\bibitem[\protect\citeauthoryear{Gruchow, Sobocinski, and Barboriak}{Gruchow
  et~al\mbox{.}}{1985}]%
        {gruchow1985alcohol}
\bibfield{author}{\bibinfo{person}{Harvey~W Gruchow},
  \bibinfo{person}{Kathleen~A Sobocinski}, {and} \bibinfo{person}{Joseph~J
  Barboriak}.} \bibinfo{year}{1985}\natexlab{}.
\newblock \showarticletitle{Alcohol, nutrient intake, and hypertension in US
  adults}.
\newblock \bibinfo{journal}{\emph{Jama}} \bibinfo{volume}{253},
  \bibinfo{number}{11} (\bibinfo{year}{1985}), \bibinfo{pages}{1567--1570}.
\newblock


\bibitem[\protect\citeauthoryear{He, Zhang, Ren, and Sun}{He
  et~al\mbox{.}}{2016}]%
        {he2016deep}
\bibfield{author}{\bibinfo{person}{Kaiming He}, \bibinfo{person}{Xiangyu
  Zhang}, \bibinfo{person}{Shaoqing Ren}, {and} \bibinfo{person}{Jian Sun}.}
  \bibinfo{year}{2016}\natexlab{}.
\newblock \showarticletitle{Deep Residual Learning for Image Recognition}. In
  \bibinfo{booktitle}{\emph{{CVPR}}}. \bibinfo{publisher}{{IEEE} Computer
  Society}, \bibinfo{pages}{770--778}.
\newblock


\bibitem[\protect\citeauthoryear{Ho, Ghosh, and Sun}{Ho et~al\mbox{.}}{2014}]%
        {ho2014marble}
\bibfield{author}{\bibinfo{person}{Joyce~C. Ho}, \bibinfo{person}{Joydeep
  Ghosh}, {and} \bibinfo{person}{Jimeng Sun}.} \bibinfo{year}{2014}\natexlab{}.
\newblock \showarticletitle{Marble: high-throughput phenotyping from electronic
  health records via sparse nonnegative tensor factorization}. In
  \bibinfo{booktitle}{\emph{{KDD}}}. \bibinfo{publisher}{{ACM}},
  \bibinfo{pages}{115--124}.
\newblock


\bibitem[\protect\citeauthoryear{Hochreiter and Schmidhuber}{Hochreiter and
  Schmidhuber}{1997}]%
        {hochreiter1997long}
\bibfield{author}{\bibinfo{person}{Sepp Hochreiter} {and}
  \bibinfo{person}{J{\"u}rgen Schmidhuber}.} \bibinfo{year}{1997}\natexlab{}.
\newblock \showarticletitle{Long short-term memory}.
\newblock \bibinfo{journal}{\emph{Neural computation}} \bibinfo{volume}{9},
  \bibinfo{number}{8} (\bibinfo{year}{1997}), \bibinfo{pages}{1735--1780}.
\newblock


\bibitem[\protect\citeauthoryear{Huang, Liu, van~der Maaten, and
  Weinberger}{Huang et~al\mbox{.}}{2017}]%
        {huang2017densely}
\bibfield{author}{\bibinfo{person}{Gao Huang}, \bibinfo{person}{Zhuang Liu},
  \bibinfo{person}{Laurens van~der Maaten}, {and} \bibinfo{person}{Kilian~Q.
  Weinberger}.} \bibinfo{year}{2017}\natexlab{}.
\newblock \showarticletitle{Densely Connected Convolutional Networks}. In
  \bibinfo{booktitle}{\emph{{CVPR}}}. \bibinfo{publisher}{{IEEE} Computer
  Society}, \bibinfo{pages}{2261--2269}.
\newblock


\bibitem[\protect\citeauthoryear{Kagaya, Aizawa, and Ogawa}{Kagaya
  et~al\mbox{.}}{2014}]%
        {kagaya2014food}
\bibfield{author}{\bibinfo{person}{Hokuto Kagaya}, \bibinfo{person}{Kiyoharu
  Aizawa}, {and} \bibinfo{person}{Makoto Ogawa}.}
  \bibinfo{year}{2014}\natexlab{}.
\newblock \showarticletitle{Food Detection and Recognition Using Convolutional
  Neural Network}. In \bibinfo{booktitle}{\emph{{ACM} Multimedia}}.
  \bibinfo{publisher}{{ACM}}, \bibinfo{pages}{1085--1088}.
\newblock


\bibitem[\protect\citeauthoryear{Kale, Che, Bahadori, Li, Liu, and Wetzel}{Kale
  et~al\mbox{.}}{2015}]%
        {kale2015causal}
\bibfield{author}{\bibinfo{person}{David~C. Kale}, \bibinfo{person}{Zhengping
  Che}, \bibinfo{person}{Mohammad~Taha Bahadori}, \bibinfo{person}{Wenzhe Li},
  \bibinfo{person}{Yan Liu}, {and} \bibinfo{person}{Randall~C. Wetzel}.}
  \bibinfo{year}{2015}\natexlab{}.
\newblock \showarticletitle{Causal Phenotype Discovery via Deep Networks}. In
  \bibinfo{booktitle}{\emph{{AMIA}}}. \bibinfo{publisher}{{AMIA}}.
\newblock


\bibitem[\protect\citeauthoryear{Kawano and Yanai}{Kawano and Yanai}{2014}]%
        {kawano2014automatic}
\bibfield{author}{\bibinfo{person}{Yoshiyuki Kawano} {and}
  \bibinfo{person}{Keiji Yanai}.} \bibinfo{year}{2014}\natexlab{}.
\newblock \showarticletitle{Automatic Expansion of a Food Image Dataset
  Leveraging Existing Categories with Domain Adaptation}. In
  \bibinfo{booktitle}{\emph{{ECCV} Workshops {(3)}}}
  \emph{(\bibinfo{series}{Lecture Notes in Computer Science},
  Vol.~\bibinfo{volume}{8927})}. \bibinfo{publisher}{Springer},
  \bibinfo{pages}{3--17}.
\newblock


\bibitem[\protect\citeauthoryear{Khosla, Teterwak, Wang, Sarna, Tian, Isola,
  Maschinot, Liu, and Krishnan}{Khosla et~al\mbox{.}}{2020}]%
        {khosla2020supervised}
\bibfield{author}{\bibinfo{person}{Prannay Khosla}, \bibinfo{person}{Piotr
  Teterwak}, \bibinfo{person}{Chen Wang}, \bibinfo{person}{Aaron Sarna},
  \bibinfo{person}{Yonglong Tian}, \bibinfo{person}{Phillip Isola},
  \bibinfo{person}{Aaron Maschinot}, \bibinfo{person}{Ce Liu}, {and}
  \bibinfo{person}{Dilip Krishnan}.} \bibinfo{year}{2020}\natexlab{}.
\newblock \showarticletitle{Supervised Contrastive Learning}. In
  \bibinfo{booktitle}{\emph{NeurIPS}}.
\newblock


\bibitem[\protect\citeauthoryear{LeCun, Bengio, and Hinton}{LeCun
  et~al\mbox{.}}{2015}]%
        {lecun2015deep}
\bibfield{author}{\bibinfo{person}{Yann LeCun}, \bibinfo{person}{Yoshua
  Bengio}, {and} \bibinfo{person}{Geoffrey~E. Hinton}.}
  \bibinfo{year}{2015}\natexlab{}.
\newblock \showarticletitle{Deep learning}.
\newblock \bibinfo{journal}{\emph{Nat.}} \bibinfo{volume}{521},
  \bibinfo{number}{7553} (\bibinfo{year}{2015}), \bibinfo{pages}{436--444}.
\newblock


\bibitem[\protect\citeauthoryear{Lipton, Kale, Elkan, and Wetzel}{Lipton
  et~al\mbox{.}}{2016}]%
        {lipton2015learning}
\bibfield{author}{\bibinfo{person}{Zachary~Chase Lipton},
  \bibinfo{person}{David~C. Kale}, \bibinfo{person}{Charles Elkan}, {and}
  \bibinfo{person}{Randall~C. Wetzel}.} \bibinfo{year}{2016}\natexlab{}.
\newblock \showarticletitle{Learning to Diagnose with {LSTM} Recurrent Neural
  Networks}. In \bibinfo{booktitle}{\emph{{ICLR} (Poster)}}.
\newblock


\bibitem[\protect\citeauthoryear{Liu, Mao, Wu, Feichtenhofer, Darrell, and
  Xie}{Liu et~al\mbox{.}}{2022}]%
        {liu2022convnet}
\bibfield{author}{\bibinfo{person}{Zhuang Liu}, \bibinfo{person}{Hanzi Mao},
  \bibinfo{person}{Chao{-}Yuan Wu}, \bibinfo{person}{Christoph Feichtenhofer},
  \bibinfo{person}{Trevor Darrell}, {and} \bibinfo{person}{Saining Xie}.}
  \bibinfo{year}{2022}\natexlab{}.
\newblock \showarticletitle{A ConvNet for the 2020s}. In
  \bibinfo{booktitle}{\emph{{CVPR}}}. \bibinfo{publisher}{{IEEE}},
  \bibinfo{pages}{11966--11976}.
\newblock


\bibitem[\protect\citeauthoryear{Ma, Chitta, Zhou, You, Sun, and Gao}{Ma
  et~al\mbox{.}}{2017}]%
        {ma2017dipole}
\bibfield{author}{\bibinfo{person}{Fenglong Ma}, \bibinfo{person}{Radha
  Chitta}, \bibinfo{person}{Jing Zhou}, \bibinfo{person}{Quanzeng You},
  \bibinfo{person}{Tong Sun}, {and} \bibinfo{person}{Jing Gao}.}
  \bibinfo{year}{2017}\natexlab{}.
\newblock \showarticletitle{Dipole: Diagnosis Prediction in Healthcare via
  Attention-based Bidirectional Recurrent Neural Networks}. In
  \bibinfo{booktitle}{\emph{{KDD}}}. \bibinfo{publisher}{{ACM}},
  \bibinfo{pages}{1903--1911}.
\newblock


\bibitem[\protect\citeauthoryear{Marr and Hildreth}{Marr and Hildreth}{1980}]%
        {marr1980theory}
\bibfield{author}{\bibinfo{person}{David Marr} {and} \bibinfo{person}{Ellen
  Hildreth}.} \bibinfo{year}{1980}\natexlab{}.
\newblock \showarticletitle{Theory of edge detection}.
\newblock \bibinfo{journal}{\emph{Proceedings of the Royal Society of London.
  Series B. Biological Sciences}} \bibinfo{volume}{207}, \bibinfo{number}{1167}
  (\bibinfo{year}{1980}), \bibinfo{pages}{187--217}.
\newblock


\bibitem[\protect\citeauthoryear{Matsuda and Yanai}{Matsuda and Yanai}{2012}]%
        {matsuda2012multiple}
\bibfield{author}{\bibinfo{person}{Yuji Matsuda} {and} \bibinfo{person}{Keiji
  Yanai}.} \bibinfo{year}{2012}\natexlab{}.
\newblock \showarticletitle{Multiple-food recognition considering co-occurrence
  employing manifold ranking}. In \bibinfo{booktitle}{\emph{{ICPR}}}.
  \bibinfo{publisher}{{IEEE} Computer Society}, \bibinfo{pages}{2017--2020}.
\newblock


\bibitem[\protect\citeauthoryear{Min, Jiang, Liu, Rui, and Jain}{Min
  et~al\mbox{.}}{2019}]%
        {min2019survey}
\bibfield{author}{\bibinfo{person}{Weiqing Min}, \bibinfo{person}{Shuqiang
  Jiang}, \bibinfo{person}{Linhu Liu}, \bibinfo{person}{Yong Rui}, {and}
  \bibinfo{person}{Ramesh~C. Jain}.} \bibinfo{year}{2019}\natexlab{}.
\newblock \showarticletitle{A Survey on Food Computing}.
\newblock \bibinfo{journal}{\emph{{ACM} Comput. Surv.}} \bibinfo{volume}{52},
  \bibinfo{number}{5} (\bibinfo{year}{2019}), \bibinfo{pages}{92:1--92:36}.
\newblock


\bibitem[\protect\citeauthoryear{Ooi, Tan, Wang, Wang, Cai, Chen, Gao, Luo,
  Tung, Wang, Xie, Zhang, and Zheng}{Ooi et~al\mbox{.}}{2015}]%
        {ooi2015singa}
\bibfield{author}{\bibinfo{person}{Beng~Chin Ooi}, \bibinfo{person}{Kian{-}Lee
  Tan}, \bibinfo{person}{Sheng Wang}, \bibinfo{person}{Wei Wang},
  \bibinfo{person}{Qingchao Cai}, \bibinfo{person}{Gang Chen},
  \bibinfo{person}{Jinyang Gao}, \bibinfo{person}{Zhaojing Luo},
  \bibinfo{person}{Anthony K.~H. Tung}, \bibinfo{person}{Yuan Wang},
  \bibinfo{person}{Zhongle Xie}, \bibinfo{person}{Meihui Zhang}, {and}
  \bibinfo{person}{Kaiping Zheng}.} \bibinfo{year}{2015}\natexlab{}.
\newblock \showarticletitle{{SINGA:} {A} Distributed Deep Learning Platform}.
  In \bibinfo{booktitle}{\emph{{ACM} Multimedia}}. \bibinfo{publisher}{{ACM}},
  \bibinfo{pages}{685--688}.
\newblock


\bibitem[\protect\citeauthoryear{Ow~Yong and Koe}{Ow~Yong and Koe}{2021}]%
        {ow2021war}
\bibfield{author}{\bibinfo{person}{Lai~Meng Ow~Yong} {and}
  \bibinfo{person}{Ling Wan~Pearline Koe}.} \bibinfo{year}{2021}\natexlab{}.
\newblock \showarticletitle{War on Diabetes in Singapore: a policy analysis}.
\newblock \bibinfo{journal}{\emph{Health Research Policy and Systems}}
  \bibinfo{volume}{19}, \bibinfo{number}{1} (\bibinfo{year}{2021}),
  \bibinfo{pages}{1--10}.
\newblock


\bibitem[\protect\citeauthoryear{Oza-Frank, Cheng, Narayan, and
  Gregg}{Oza-Frank et~al\mbox{.}}{2009}]%
        {oza2009trends}
\bibfield{author}{\bibinfo{person}{Reena Oza-Frank}, \bibinfo{person}{Yiling~J
  Cheng}, \bibinfo{person}{KM~Venkat Narayan}, {and} \bibinfo{person}{Edward~W
  Gregg}.} \bibinfo{year}{2009}\natexlab{}.
\newblock \showarticletitle{Trends in nutrient intake among adults with
  diabetes in the United States: 1988-2004}.
\newblock \bibinfo{journal}{\emph{Journal of the American Dietetic
  Association}} \bibinfo{volume}{109}, \bibinfo{number}{7}
  (\bibinfo{year}{2009}), \bibinfo{pages}{1173--1178}.
\newblock


\bibitem[\protect\citeauthoryear{Sandler, Howard, Zhu, Zhmoginov, and
  Chen}{Sandler et~al\mbox{.}}{2018}]%
        {sandler2018mobilenetv2}
\bibfield{author}{\bibinfo{person}{Mark Sandler}, \bibinfo{person}{Andrew~G.
  Howard}, \bibinfo{person}{Menglong Zhu}, \bibinfo{person}{Andrey Zhmoginov},
  {and} \bibinfo{person}{Liang{-}Chieh Chen}.} \bibinfo{year}{2018}\natexlab{}.
\newblock \showarticletitle{MobileNetV2: Inverted Residuals and Linear
  Bottlenecks}. In \bibinfo{booktitle}{\emph{{CVPR}}}.
  \bibinfo{publisher}{Computer Vision Foundation / {IEEE} Computer Society},
  \bibinfo{pages}{4510--4520}.
\newblock


\bibitem[\protect\citeauthoryear{Sutskever, Martens, Dahl, and
  Hinton}{Sutskever et~al\mbox{.}}{2013}]%
        {sutskever2013on}
\bibfield{author}{\bibinfo{person}{Ilya Sutskever}, \bibinfo{person}{James
  Martens}, \bibinfo{person}{George~E. Dahl}, {and}
  \bibinfo{person}{Geoffrey~E. Hinton}.} \bibinfo{year}{2013}\natexlab{}.
\newblock \showarticletitle{On the importance of initialization and momentum in
  deep learning}. In \bibinfo{booktitle}{\emph{{ICML} {(3)}}}
  \emph{(\bibinfo{series}{{JMLR} Workshop and Conference Proceedings},
  Vol.~\bibinfo{volume}{28})}. \bibinfo{publisher}{JMLR.org},
  \bibinfo{pages}{1139--1147}.
\newblock


\bibitem[\protect\citeauthoryear{Tan and Le}{Tan and Le}{2021}]%
        {tan2021efficientnetv2}
\bibfield{author}{\bibinfo{person}{Mingxing Tan} {and} \bibinfo{person}{Quoc~V.
  Le}.} \bibinfo{year}{2021}\natexlab{}.
\newblock \showarticletitle{EfficientNetV2: Smaller Models and Faster
  Training}. In \bibinfo{booktitle}{\emph{{ICML}}}
  \emph{(\bibinfo{series}{Proceedings of Machine Learning Research},
  Vol.~\bibinfo{volume}{139})}. \bibinfo{publisher}{{PMLR}},
  \bibinfo{pages}{10096--10106}.
\newblock


\bibitem[\protect\citeauthoryear{Tan and Arcaya}{Tan and Arcaya}{2020}]%
        {tan2020we}
\bibfield{author}{\bibinfo{person}{Shin~Bin Tan} {and} \bibinfo{person}{Mariana
  Arcaya}.} \bibinfo{year}{2020}\natexlab{}.
\newblock \showarticletitle{Where we eat is who we are: a survey of
  food-related travel patterns to Singapore’s hawker centers, food courts and
  coffee shops}.
\newblock \bibinfo{journal}{\emph{International Journal of Behavioral Nutrition
  and Physical Activity}} \bibinfo{volume}{17}, \bibinfo{number}{1}
  (\bibinfo{year}{2020}), \bibinfo{pages}{1--14}.
\newblock


\bibitem[\protect\citeauthoryear{Thompson, Subar, Loria, Reedy, and
  Baranowski}{Thompson et~al\mbox{.}}{2010}]%
        {thompson2010need}
\bibfield{author}{\bibinfo{person}{Frances~E Thompson}, \bibinfo{person}{Amy~F
  Subar}, \bibinfo{person}{Catherine~M Loria}, \bibinfo{person}{Jill~L Reedy},
  {and} \bibinfo{person}{Tom Baranowski}.} \bibinfo{year}{2010}\natexlab{}.
\newblock \showarticletitle{Need for technological innovation in dietary
  assessment}.
\newblock \bibinfo{journal}{\emph{Journal of the American Dietetic
  Association}} \bibinfo{volume}{110}, \bibinfo{number}{1}
  (\bibinfo{year}{2010}), \bibinfo{pages}{48}.
\newblock


\bibitem[\protect\citeauthoryear{Tucker}{Tucker}{2016}]%
        {tucker2016nutrient}
\bibfield{author}{\bibinfo{person}{Katherine~L Tucker}.}
  \bibinfo{year}{2016}\natexlab{}.
\newblock \showarticletitle{Nutrient intake, nutritional status, and cognitive
  function with aging}.
\newblock \bibinfo{journal}{\emph{Annals of the New York Academy of Sciences}}
  \bibinfo{volume}{1367}, \bibinfo{number}{1} (\bibinfo{year}{2016}),
  \bibinfo{pages}{38--49}.
\newblock


\bibitem[\protect\citeauthoryear{Wang, Zhang, Chen, Jagadish, Ooi, and
  Tan}{Wang et~al\mbox{.}}{2016}]%
        {wang2016database}
\bibfield{author}{\bibinfo{person}{Wei Wang}, \bibinfo{person}{Meihui Zhang},
  \bibinfo{person}{Gang Chen}, \bibinfo{person}{H.~V. Jagadish},
  \bibinfo{person}{Beng~Chin Ooi}, {and} \bibinfo{person}{Kian{-}Lee Tan}.}
  \bibinfo{year}{2016}\natexlab{}.
\newblock \showarticletitle{Database Meets Deep Learning: Challenges and
  Opportunities}.
\newblock \bibinfo{journal}{\emph{{SIGMOD} Rec.}} \bibinfo{volume}{45},
  \bibinfo{number}{2} (\bibinfo{year}{2016}), \bibinfo{pages}{17--22}.
\newblock


\bibitem[\protect\citeauthoryear{Wang, Sontag, and Wang}{Wang
  et~al\mbox{.}}{2014}]%
        {wang2014unsupervised}
\bibfield{author}{\bibinfo{person}{Xiang Wang}, \bibinfo{person}{David~A.
  Sontag}, {and} \bibinfo{person}{Fei Wang}.} \bibinfo{year}{2014}\natexlab{}.
\newblock \showarticletitle{Unsupervised learning of disease progression
  models}. In \bibinfo{booktitle}{\emph{{KDD}}}. \bibinfo{publisher}{{ACM}},
  \bibinfo{pages}{85--94}.
\newblock


\bibitem[\protect\citeauthoryear{Wu, Merler, Uceda{-}Sosa, and Smith}{Wu
  et~al\mbox{.}}{2016}]%
        {wu2016learning}
\bibfield{author}{\bibinfo{person}{Hui Wu}, \bibinfo{person}{Michele Merler},
  \bibinfo{person}{Rosario Uceda{-}Sosa}, {and} \bibinfo{person}{John~R.
  Smith}.} \bibinfo{year}{2016}\natexlab{}.
\newblock \showarticletitle{Learning to Make Better Mistakes: Semantics-aware
  Visual Food Recognition}. In \bibinfo{booktitle}{\emph{{ACM} Multimedia}}.
  \bibinfo{publisher}{{ACM}}, \bibinfo{pages}{172--176}.
\newblock


\bibitem[\protect\citeauthoryear{Yang, Chen, Pomerleau, and Sukthankar}{Yang
  et~al\mbox{.}}{2010}]%
        {yang2010food}
\bibfield{author}{\bibinfo{person}{Shulin Yang}, \bibinfo{person}{Mei Chen},
  \bibinfo{person}{Dean Pomerleau}, {and} \bibinfo{person}{Rahul Sukthankar}.}
  \bibinfo{year}{2010}\natexlab{}.
\newblock \showarticletitle{Food recognition using statistics of pairwise local
  features}. In \bibinfo{booktitle}{\emph{{CVPR}}}. \bibinfo{publisher}{{IEEE}
  Computer Society}, \bibinfo{pages}{2249--2256}.
\newblock


\bibitem[\protect\citeauthoryear{Yosinski, Clune, Bengio, and Lipson}{Yosinski
  et~al\mbox{.}}{2014}]%
        {yosinski2014how}
\bibfield{author}{\bibinfo{person}{Jason Yosinski}, \bibinfo{person}{Jeff
  Clune}, \bibinfo{person}{Yoshua Bengio}, {and} \bibinfo{person}{Hod Lipson}.}
  \bibinfo{year}{2014}\natexlab{}.
\newblock \showarticletitle{How transferable are features in deep neural
  networks?}. In \bibinfo{booktitle}{\emph{{NIPS}}}.
  \bibinfo{pages}{3320--3328}.
\newblock


\bibitem[\protect\citeauthoryear{Zhang, Gao, Ma, Wang, Wang, and Tang}{Zhang
  et~al\mbox{.}}{2021}]%
        {zhang2021grasp}
\bibfield{author}{\bibinfo{person}{Chaohe Zhang}, \bibinfo{person}{Xin Gao},
  \bibinfo{person}{Liantao Ma}, \bibinfo{person}{Yasha Wang},
  \bibinfo{person}{Jiangtao Wang}, {and} \bibinfo{person}{Wen Tang}.}
  \bibinfo{year}{2021}\natexlab{}.
\newblock \showarticletitle{{GRASP:} Generic Framework for Health Status
  Representation Learning Based on Incorporating Knowledge from Similar
  Patients}. In \bibinfo{booktitle}{\emph{{AAAI}}}. \bibinfo{publisher}{{AAAI}
  Press}, \bibinfo{pages}{715--723}.
\newblock


\bibitem[\protect\citeauthoryear{Zhang, Isola, Efros, Shechtman, and
  Wang}{Zhang et~al\mbox{.}}{2018}]%
        {zhang2018unreasonable}
\bibfield{author}{\bibinfo{person}{Richard Zhang}, \bibinfo{person}{Phillip
  Isola}, \bibinfo{person}{Alexei~A. Efros}, \bibinfo{person}{Eli Shechtman},
  {and} \bibinfo{person}{Oliver Wang}.} \bibinfo{year}{2018}\natexlab{}.
\newblock \showarticletitle{The Unreasonable Effectiveness of Deep Features as
  a Perceptual Metric}. In \bibinfo{booktitle}{\emph{{CVPR}}}.
  \bibinfo{publisher}{Computer Vision Foundation / {IEEE} Computer Society},
  \bibinfo{pages}{586--595}.
\newblock


\bibitem[\protect\citeauthoryear{Zheng, Cai, Chua, Herschel, Zhang, and
  Ooi}{Zheng et~al\mbox{.}}{2022a}]%
        {zheng2022dyhealth}
\bibfield{author}{\bibinfo{person}{Kaiping Zheng}, \bibinfo{person}{Shaofeng
  Cai}, \bibinfo{person}{Horng~Ruey Chua}, \bibinfo{person}{Melanie Herschel},
  \bibinfo{person}{Meihui Zhang}, {and} \bibinfo{person}{Beng~Chin Ooi}.}
  \bibinfo{year}{2022}\natexlab{a}.
\newblock \showarticletitle{DyHealth: Making Neural Networks Dynamic for
  Effective Healthcare Analytics}.
\newblock \bibinfo{journal}{\emph{Proc. {VLDB} Endow.}} \bibinfo{volume}{15},
  \bibinfo{number}{12} (\bibinfo{year}{2022}), \bibinfo{pages}{3445--3458}.
\newblock


\bibitem[\protect\citeauthoryear{Zheng, Cai, Chua, Wang, Ngiam, and Ooi}{Zheng
  et~al\mbox{.}}{2020}]%
        {zheng2020tracer}
\bibfield{author}{\bibinfo{person}{Kaiping Zheng}, \bibinfo{person}{Shaofeng
  Cai}, \bibinfo{person}{Horng~Ruey Chua}, \bibinfo{person}{Wei Wang},
  \bibinfo{person}{Kee~Yuan Ngiam}, {and} \bibinfo{person}{Beng~Chin Ooi}.}
  \bibinfo{year}{2020}\natexlab{}.
\newblock \showarticletitle{{TRACER:} {A} Framework for Facilitating Accurate
  and Interpretable Analytics for High Stakes Applications}. In
  \bibinfo{booktitle}{\emph{{SIGMOD} Conference}}. \bibinfo{publisher}{{ACM}},
  \bibinfo{pages}{1747--1763}.
\newblock


\bibitem[\protect\citeauthoryear{Zheng, Chen, Herschel, Ngiam, Ooi, and
  Gao}{Zheng et~al\mbox{.}}{2021}]%
        {zheng2021pace}
\bibfield{author}{\bibinfo{person}{Kaiping Zheng}, \bibinfo{person}{Gang Chen},
  \bibinfo{person}{Melanie Herschel}, \bibinfo{person}{Kee~Yuan Ngiam},
  \bibinfo{person}{Beng~Chin Ooi}, {and} \bibinfo{person}{Jinyang Gao}.}
  \bibinfo{year}{2021}\natexlab{}.
\newblock \showarticletitle{{PACE:} Learning Effective Task Decomposition for
  Human-in-the-loop Healthcare Delivery}. In \bibinfo{booktitle}{\emph{{SIGMOD}
  Conference}}. \bibinfo{publisher}{{ACM}}, \bibinfo{pages}{2156--2168}.
\newblock


\bibitem[\protect\citeauthoryear{Zheng, Gao, Ngiam, Ooi, and Yip}{Zheng
  et~al\mbox{.}}{2017a}]%
        {zheng2017resolving}
\bibfield{author}{\bibinfo{person}{Kaiping Zheng}, \bibinfo{person}{Jinyang
  Gao}, \bibinfo{person}{Kee~Yuan Ngiam}, \bibinfo{person}{Beng~Chin Ooi},
  {and} \bibinfo{person}{James Wei~Luen Yip}.}
  \bibinfo{year}{2017}\natexlab{a}.
\newblock \showarticletitle{Resolving the Bias in Electronic Medical Records}.
  In \bibinfo{booktitle}{\emph{{KDD}}}. \bibinfo{publisher}{{ACM}},
  \bibinfo{pages}{2171--2180}.
\newblock


\bibitem[\protect\citeauthoryear{Zheng, Nguyen, Liu, Goh, and Ooi}{Zheng
  et~al\mbox{.}}{2022b}]%
        {zheng2022edental}
\bibfield{author}{\bibinfo{person}{Kaiping Zheng}, \bibinfo{person}{Thao
  Nguyen}, \bibinfo{person}{Changshuo Liu}, \bibinfo{person}{Charlene~Enhui
  Goh}, {and} \bibinfo{person}{Beng~Chin Ooi}.}
  \bibinfo{year}{2022}\natexlab{b}.
\newblock \showarticletitle{eDental: Managing Your Dental Care in Diet
  Diaries}. In \bibinfo{booktitle}{\emph{{CIKM}}}. \bibinfo{publisher}{{ACM}},
  \bibinfo{pages}{5059--5063}.
\newblock


\bibitem[\protect\citeauthoryear{Zheng, Wang, Gao, Ngiam, Ooi, and Yip}{Zheng
  et~al\mbox{.}}{2017b}]%
        {zheng2017capturing}
\bibfield{author}{\bibinfo{person}{Kaiping Zheng}, \bibinfo{person}{Wei Wang},
  \bibinfo{person}{Jinyang Gao}, \bibinfo{person}{Kee~Yuan Ngiam},
  \bibinfo{person}{Beng~Chin Ooi}, {and} \bibinfo{person}{James Wei~Luen Yip}.}
  \bibinfo{year}{2017}\natexlab{b}.
\newblock \showarticletitle{Capturing Feature-Level Irregularity in Disease
  Progression Modeling}. In \bibinfo{booktitle}{\emph{{CIKM}}}.
  \bibinfo{publisher}{{ACM}}, \bibinfo{pages}{1579--1588}.
\newblock


\bibitem[\protect\citeauthoryear{Zhu, Luo, Wang, Zhang, Chen, and Zheng}{Zhu
  et~al\mbox{.}}{2021}]%
        {zhu2021towards}
\bibfield{author}{\bibinfo{person}{Lei Zhu}, \bibinfo{person}{Zhaojing Luo},
  \bibinfo{person}{Wei Wang}, \bibinfo{person}{Meihui Zhang},
  \bibinfo{person}{Gang Chen}, {and} \bibinfo{person}{Kaiping Zheng}.}
  \bibinfo{year}{2021}\natexlab{}.
\newblock \showarticletitle{Towards Robust Cross-domain Image Understanding
  with Unsupervised Noise Removal}. In \bibinfo{booktitle}{\emph{{ACM}
  Multimedia}}. \bibinfo{publisher}{{ACM}}, \bibinfo{pages}{3024--3033}.
\newblock


\end{thebibliography}

\end{document}